\documentclass[journal,twoside]{IEEEtran}

\ifCLASSINFOpdf
  \usepackage[pdftex]{graphicx}
  \graphicspath{{figures/}}
  \DeclareGraphicsExtensions{.pdf,.jpeg,.png}
\else
  \usepackage[dvips]{graphicx}
  \graphicspath{{figures/}}
  \DeclareGraphicsExtensions{.eps}
\fi

\usepackage{diagbox}
\usepackage{multirow}

\usepackage{cite}

\usepackage{color,xcolor}

\usepackage{amsmath}
\usepackage{amsfonts}

\usepackage{array}

\ifCLASSOPTIONcompsoc
  \usepackage[caption=false,font=normalsize,labelfont=sf,textfont=sf]{subfig}
\else
  \usepackage[caption=false,font=footnotesize]{subfig}
\fi
\usepackage{float}

\usepackage{threeparttable}

\usepackage{lipsum}

\usepackage{url}

\begin{document}
\title{Spatio-Temporal Feedback Control of Small Target Motion Detection Visual System}
\author{Hongxin Wang, Zhiyan Zhong, Fang Lei, Xiaohua Jing, Jigen Peng, and Shigang Yue, \IEEEmembership{Senior Member,~IEEE}  \thanks{This work was supported in part by the National Natural Science Foundation of China under Grant 12031003 and Grant 62103112, in part by the European Union’s Horizon 2020 research and innovation programme under the Marie Sklodowska-Curie grant agreement No 691154 STEP2DYNA and No 778062 ULTRACEPT, in part by the China Postdoctoral Science Foundation under Grant 2021M700921. \emph{(Corresponding authors:  Shigang Yue \& Jigen Peng.)}}
\thanks{Hongxin Wang, Zhiyan Zhong, Fang Lei, Xiaohua Jing, and Shigang Yue are with the Computational Intelligence Lab, School of Computer Science, University of Lincoln, Lincoln LN6 7TS, U.K. (email: howang@lincoln.ac.uk, syue@lincoln.ac.uk).}
\thanks{Jigen Peng is with the School of Mathematics and Information Science, Guangzhou University, Guangzhou 510006, China (email: jgpeng@gzhu.edu.cn).}
}
   

\markboth{IEEE TRANSACTIONS ON Neural Networks and Learning Systems}
{Wang \MakeLowercase{\textit{et al.}}: Spatio-Temporal Feedback Control of Small Target Motion Detection Neural Network}
%



\maketitle

\begin{abstract}
Feedback is crucial to motion perception in animals' visual systems where its spatial and temporal dynamics are often shaped by movement patterns of surrounding environments. However, such spatio-temporal feedback has not been deeply explored in designing neural networks to detect small moving targets that cover only one or a few pixels in image while presenting extremely limited visual features. In this paper, we address small target motion detection problem by developing a visual system with spatio-temporal feedback loop, and further reveal its important roles in suppressing false positive background movement while enhancing network responses to small targets. Specifically, the proposed visual system is composed of two complementary subnetworks. The first subnetwork is designed to extract spatial and temporal motion patterns of cluttered backgrounds by neuronal ensemble coding. The second subnetwork is developed to capture small target motion information and integrate the spatio-temporal feedback signal from the first subnetwork to inhibit background false positives. Experimental results demonstrate that the proposed spatio-temporal feedback visual system is more competitive than existing methods in discriminating small moving targets from complex dynamic environment.
\end{abstract}


%
\IEEEpeerreviewmaketitle

\section{Introduction}
\label{Introduction}

\begin{figure}[!t]
	\centering
	\includegraphics[width=0.385\textwidth]{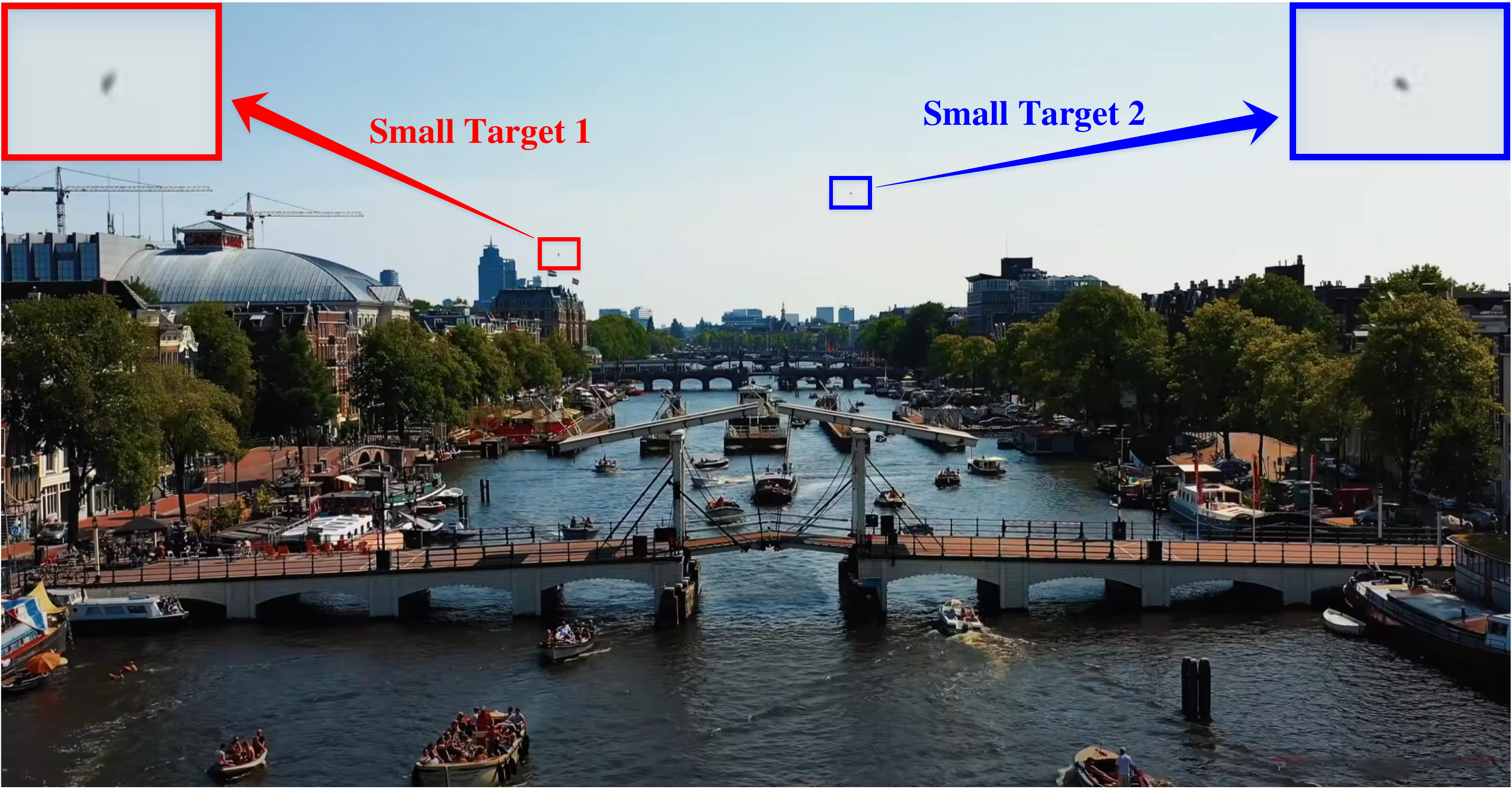}
	\caption{Two small targets in the far distance that could be bird or drone \cite{Youtube-Small-Objects}. The red-boxed and blue-boxed regions are enlarged, each of which contains a small target and its surrounding background. Due to long observation distance, the targets all appear as minute dim speckles in image, only a few pixels in size, presenting extremely weak visual features.}
	\label{Examples-of-Small-Targets}	
\end{figure}

Autonomous mobile robots with visual sensors onboard, such as drones, space probes, and unmanned underwater vehicles, have shown great potential in performing a wide range of challenging tasks, from navigation in uncontrolled environments to military reconnaissance without the need for guidance devices \cite{yue2006collision,webb2020robots}. Artificial visual systems capable of efficient and robust motion perception are important for mobile robots to enable autonomic responses to surroundings that is always highly complex, dynamic, and abundant with visual motion. For example, early detection of objects with potential threats in the distance would help intelligent robots readily seize a first-mover advantage in competition or interaction. However, when a target is far distant or extremely small, it generally cover only one or a few pixels in image, appearing as a dim speckle without prominent appearance information. Fig. \ref{Examples-of-Small-Targets} gives an example of two small moving targets in the distance. Due to their limited sizes and few appearance cues, it is difficult even for humans to perceive such tiny objects.

Small target motion detection against complex dynamic environment finds applications in a variety of domains including autonomous driving, military reconnaissance, and astronomic observation. However, it remains highly challenging to artificial visual systems, because 1) small targets are only one or a few pixels in size, exhibiting poor-quality appearance, let alone process discriminative visual features for motion detection; 2) small targets always display highly blurred boundaries and extremely low contrast against complex backgrounds, making them difficult to distinguish from noisy background clutters; 3) freely moving camera could introduce further challenges into motion discrimination, such as strong parallax effects, constantly changing complex scene, and significant relative motion to small targets.

Great progress has been made towards motion detection of large objects with sufficiently detailed appearance and clearly defined structure from which rich visual features could be extracted. Traditionally, object motion detection is  approached in motion-based methods or appearance-based methods. The former relies primarily on luminance changes of each pixel over time to estimate object motion and can be further subdivided into background subtraction \cite{garcia2020background}, temporal differencing \cite{xiao2010vehicle}, and optical flow \cite{fortun2015optical}. The later focuses on utilizing machine learning algorithms, for example, convolutional neural networks \cite{song2018fast}, to learn visual features of objects in individual image and then match the extracted features in consecutive images to infer object movement. However, these conventional methods would always suffer from two major problems when directly applied to the task of small target motion detection. First, distinctive visual feature representations are extremely difficult to extract/learn from poor-quality appearances of small objects that are only one or a few pixels in size. In addition, background variations induced by mobile cameras highly degenerate their performance, since minute errors resulting form frame registration easily submerge small targets and lead to a number of false-positive detections. Consequently, effective solutions are required to bridge the performance gap in small target motion detection.






Despite their low-resolution eyes and tiny brains, insects provide an elegant solution to discern small moving targets against complex dynamic environment robustly with limited computational resources \cite{nordstrom2009feature}. For example, dragonflies are quite apt at chasing small mates or prey while performing sophisticated fast aerial maneuvers, evidenced by extremely high successful capture rate over $95\%$ \cite{bekkouche2021modeling}. A group of neuron subtypes in insects' visual systems, called small target motion detectors (STMDs), are believed to underlie such excellent sensitivity \cite{nordstrom2006insect,nordstrom2012neural,fabian2021spike}. They are strongly excited by movement of small targets subtending between $1^{\circ}$ and $3^{\circ}$ of the visual field, whereas moving objects subtending larger than $10^{\circ}$ always elicit much weaker neural responses. Moreover, the STMD neurons respond robustly to small target motion even in complex dynamic environment, giving neither excitatory nor inhibitory responses to background clutters. Learning from insects' visual system and its neural implementation is clearly a promising way forward in building robust and efficient machine intelligence for small target motion detection.

To reproduce the superior properties of the STMD neurons in practical applications, considerable efforts have been undertaken to establish explicit correspondence between insects' neural circuits and artificial visual systems. For example, an elementary STMD (ESTMD) model \cite{wiederman2008model} was proposed to describe the remarkable selectivity of the STMD neurons for object sizes. To account for direction selectivity, cascaded models \cite{wiederman2013biologically} and directionally selective STMD (DSTMD) \cite{wang2018directionally} were successively developed. Combining the DSTMD with a contrast pathway in parallel, STMD plus \cite{wang2019Robust} was designed to reveal the effect of information fusion on motion discrimination. These STMD-based models are all composed of multiple neural layers interconnected in a feed-forward way to process visual stimuli sequentially. Although these models can partly explain the underlying neural computation of the STMD, their outputs often contain numerous background false positives from which real small target motion is difficult to be separated. The robustness of these models needs to be improved in dealing with small target detection against complex natural background. To suppress false positive background motion, time-delay feedback connection was introduced into the STMD-based methods \cite{wang2021time}. However, the time-delay feedback STMD makes certain assumption that small targets should be faster than background, which means it can only inhibit those slow-moving background false positives.

To solve the above problems, we propose a STMD-based visual system with spatio-temporal feedback mechanism. In insects' visual systems, motion patterns of surroundings are perceivable as spatial and temporal variations of brightness on the retina, which in turn shape dynamics of feedback signals \cite{paredes2019unsupervised,stamper2012active,layton2014neural}. To enhance object motion of interest while suppressing distracting signals from cluttered background, spatio-temporal feedback is desired but has not been deeply explored \cite{zhao2020soft}. The proposed visual system is mainly composed of two complementary subnetworks, where the first one is designed for inferring spatio-temporal dynamics of cluttered background while the second one is developed for capturing cues of small target movement. We devise a feedback connection between the two subnetworks, where the spatio-temporal information about background movement is integrated with the output of the second subnetwork to suppress background false positives in self-recurrent manner. 

We organize the rest of this paper as follows. Section \ref{Related-Work} briefly overviews the related works. Section \ref{Spatio-Temporal-Feedback-Visual-System} introduces the details of the spatio-temporal feedback visual system. Section \ref{Experiments} reports extensive experiment results together with qualitative studies on synthetic and real-world datasets. Finally, Section \ref{Conclusion} provides concluding remarks.

\section{Related Work}
\label{Related-Work}

\subsection{Neuroscience-Inspired Motion Perception}
Inspiration from neuroscience is a promising approach for designing artificial visual systems with requirement of a high level of efficiency and robustness but limited in computational and memory budget \cite{webb2020robots,de2022insect}. It has attracted a great deal of interests and become an emerging research area with a number of practical applications, such as visually guided flights or landing \cite{wang2020bioinspired,mischiati2015internal}, autonomous navigation \cite{sun2020decentralised,xiong2021no}, and collision detection \cite{zhao2021enhancing,fu2019robust,Fang9718016Looming}. Our work is mainly related to two types of widely investigated motion-sensitive neurons, called lobula plate tangential cells (LPTCs) \cite{TUTHILL2016677,henning2022populations} and small target motion detectors (STMDs), whose biological properties and computational models are briefly discussed. 

LPTC neurons found in lobula plate of insects' visual systems, exhibit strong preference to object motion occupying large parts of the visual field (called wide-field motion). It is initially modelled by an array of elementary motion detectors (EMDs) \cite{hassenstein1956systemtheoretische}, each of which perceives object motion in a small part of the visual field. Specifically, a single EMD model relies on multiplication of luminance-change signals from two neighboring pixels to generate positive outputs to object movement in its preferred direction, one of which have been time-delayed for signal alignment in temporal domain. Furthermore, two EMDs can be combined in a mirror-symmetric manner to discriminate responses to objects moving along preferred direction and null direction \cite{iida2000navigation}. Biological research revealed that luminance increase and decrease signals are processed in parallel by medulla neural pathways and then recombined by the LPTC neurons for motion detection \cite{behnia2014processing}. Taking this new finding into account, researchers proposed to split the input luminance-change signal into two parallel channels, which encode luminance increments (ON) and decrements (OFF), respectively. These two channels are then combined in possible pairs, giving rise to several variants of the EMD, such as Weighted Quadrant Detector \cite{clark2011defining}, Two Quadrant Detector \cite{eichner2011internal}, and Four Quadrant Detector \cite{eichner2011internal}.

STMD neurons are highly sensitive to movement of small objects that occupy $1^{\circ}–3^{\circ}$ of the visual field. Inspired by their physiological characteristics, there have been many attempts to design the STMD-based neural networks for small target motion detection. For example, the elementary STMD  (ESTMD) \cite{wiederman2008model} model was proposed to identify the presence of small moving objects by implementing lateral inhibition and signal correlation mechanisms on intensity changes at each pixel. Specifically, it first separates the signal of intensity changes at each pixel into increase and decrease parts using half-wave rectification, which are further laterally inhibited to suppress large object motion. After that, the increase component is correlated with the time-delay decrease component via a multiplying unit to simulate the STMD responses. The ESTMD model was later improved by Wiederman \emph{et al.} \cite{wiederman2013biologically} to reproduce direction selectivity of the STMD neurons. They proposed to cascade the ESTMD with the EMD \cite{hassenstein1956systemtheoretische}, resulting in two directionally selective models, called EMD-ESTMD and ESTMD-EMD. Wang \emph{et al.}  \cite{wang2018directionally} provided an alternative called DSTMD to generate direction selectivity by correlating luminance-change signals from two different pixels. Based on insects' multi-information fusion schemes, the STMD plus model \cite{wang2019Robust} was developed by parallelly integrating spatial contrast with motion information to eliminate background features that are highly similar to small targets. However, the aforementioned models are all characterized by a feedforward hierarchy to process visual stimuli via multiple sequentially arranged neural layers. Although these feedforward models exhibit parts of selectivities the same as the STMD neurons, they generally yield numerous false positives in dealing with small target detection against complex dynamic environment. To suppress false positive background movement, a time-delay feedback mechanism was incorporated into the STMD-based neural network \cite{wang2021time}. However, the time-delay feedback can only filter out slow-moving background motion. That is, it would be powerless against those false positives with high speed.

\subsection{Spatio-Temporal Feedback}
Motion perception in insects' visual systems involves propagation and transformation of visual information across multiple hierarchically organized neural layers each of which functionally specializes for processing various aspects of object motion \cite{kwon2016sensory}. These neural layers interact closely in both feedforward and feedback directions, where feedforward conveys visual signals to higher layers while feedback passes high-level semantic information down to lower layers for modulating neural coding, removing distracting signals, and optimizing motion estimation \cite{clarke2017feedback}. 

When an insect performs high-speed maneuvers against complex dynamic environments, its self-motion relative to surroundings casts spatio-temporal variations of brightness on the retina, making small target motion discrimination a huge challenge. However, insects still demonstrate remarkable abilities to robustly discern and tracks moving objects of interest, such as predators, prey, and conspecifics, against cluttered backgrounds. Biological research indicates that motion patterns of surroundings would shape spatial and temporal dynamics of feedback signals which in turn suppress neural responses to distracting background objects while enhancing those to small moving targets \cite{paredes2019unsupervised,stamper2012active,layton2014neural}. Moreover, such spatio-temporal feedback is propagated from the lobula plate tangential cells (LPTCs) to target-selective neurons for filtering out background motion \cite{mauss2015neural,fenk2014asymmetric,nicholas2021facilitation}. 

Feedback has been focus of studies on network structure and proved to be capable of significantly boosting network performances in various tasks, such as visual segmentation \cite{cao2018feedback}, saliency detection \cite{li2018contrast}, and object recognition \cite{han2018deep}. Despite its success in these computer vision applications, feedback with specific spatio-temporal dynamics has not been incorporated in the STMD-based neural networks and its functional significance remains unclear.


\section{Spatio-Temporal Feedback Visual System}
\label{Spatio-Temporal-Feedback-Visual-System}

\begin{figure}[!t]
	\centering
	\includegraphics[width=0.48\textwidth]{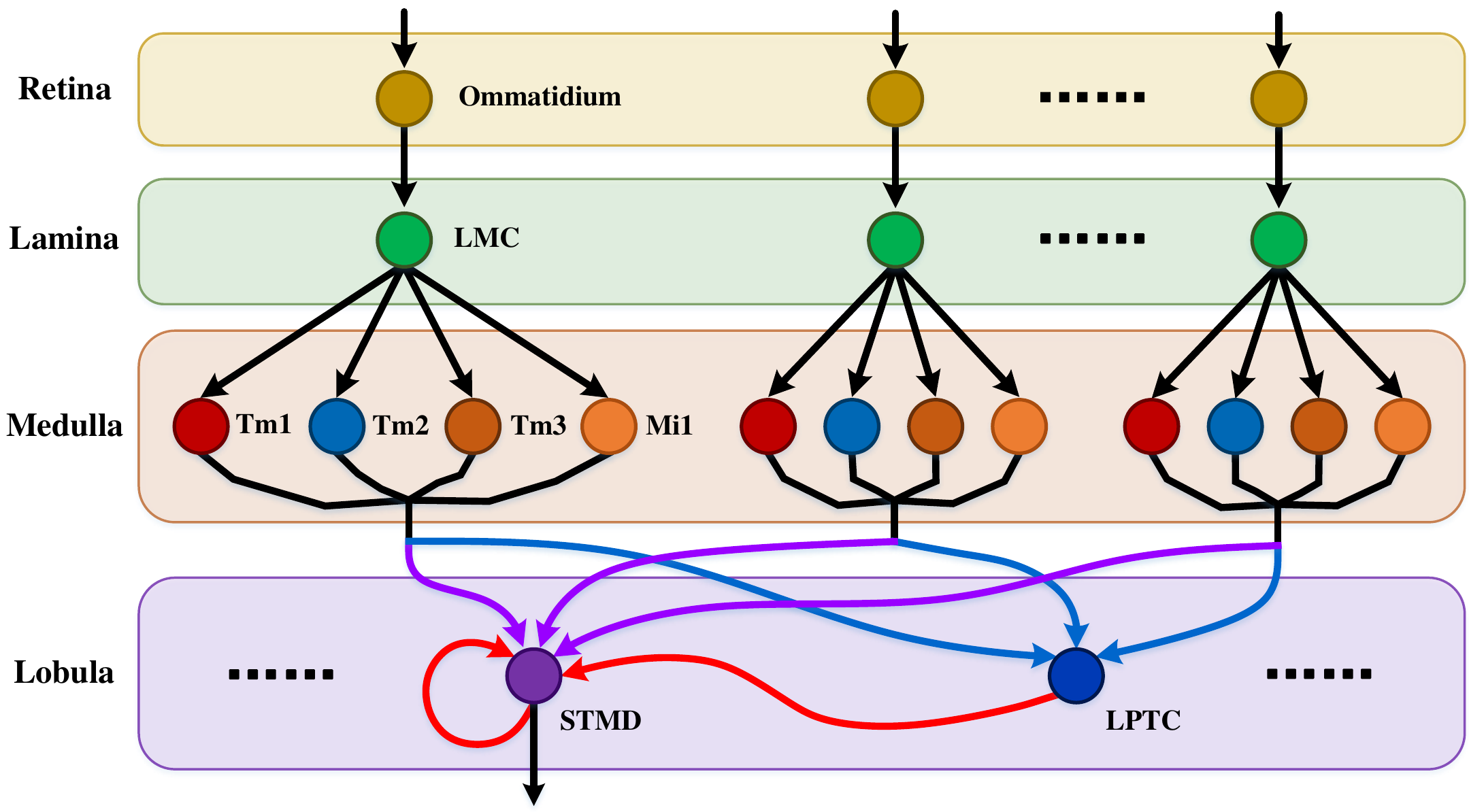}
	\caption{Network structure of the proposed spatio-temporal feedback visual system. It is hierarchically organized and composed of four sequentially arranged neural layers, including retina, lamina, medulla, and lobula (from top to bottom), each of which comprises numerous specialized neurons denoted by colored nodes. Red lines denote feedback loops while other colored lines represent feedforward connections.}
	\label{Wiring-Sketch-Spatio-Temporal-Feedback}
\end{figure}


The proposed spatio-temporal feedback visual system is a multi-layer network with a number of specialized neurons, as shown in Fig.  \ref{Wiring-Sketch-Spatio-Temporal-Feedback}. External visual stimuli successively flow through ommatidia in retina layer \cite{caves2018visual}, large monopolar cells (LMCs) in lamina layer \cite{stockl2020hawkmoth},  medulla neurons (e.g., Tm1, Tm2, Tm3, and Mi1) in medulla layer \cite{shinomiya2019comparisons}, and are integrated by the STMD \cite{fabian2021spike} and LPTC \cite{henning2022populations} neurons in lobula layer, respectively. Besides the feedforward connections, the STMD neuron applies its output to its input in a recurrent manner, forming a self-feedback connection, while the LPTC neuron propagates its output incorporating spatio-temporal environmental dynamics to the input of the STMD neuron through an intra-layer feedback loop. We describe functionalities of each neural layer and their formulations in Sections \ref{Retina-Layer}--\ref{Lobula-Layer}.



\begin{figure*}[!t]
	\centering
	\includegraphics[width=0.90\textwidth]{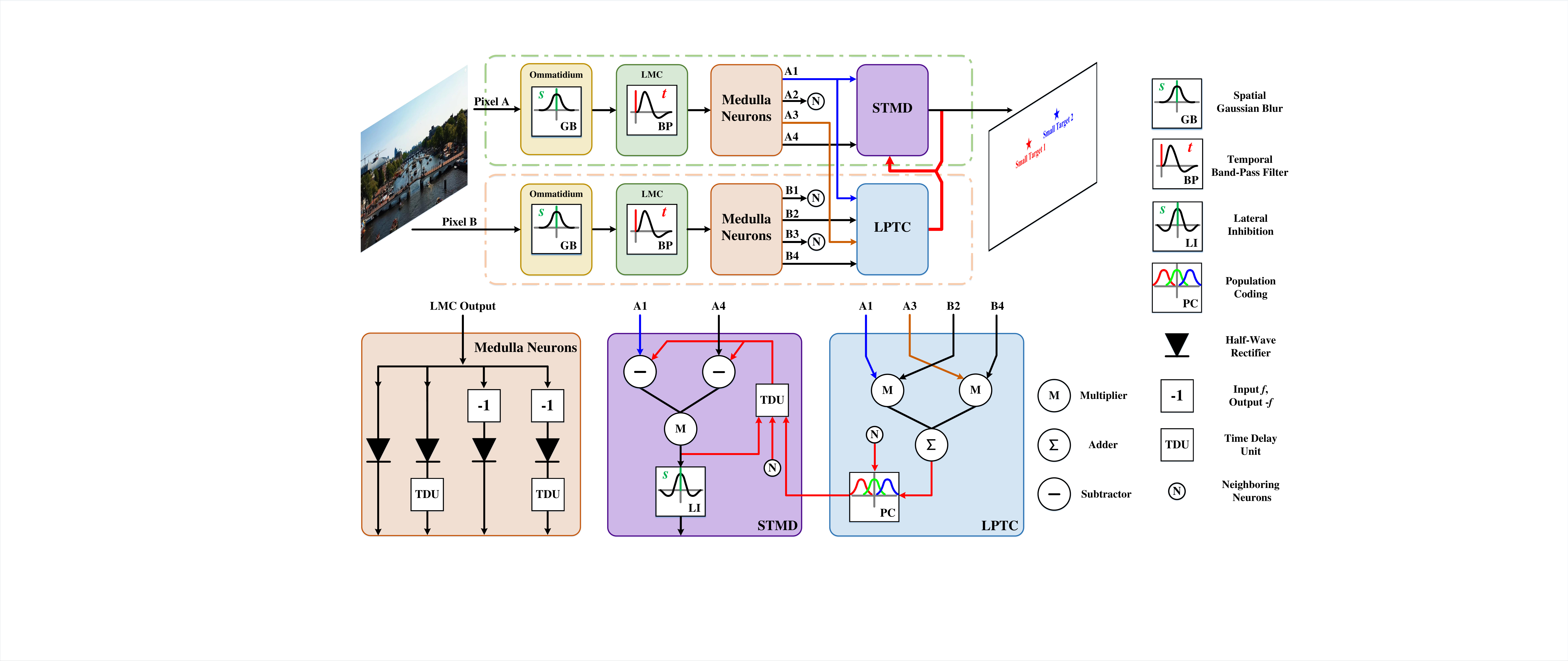}
	\caption{Schematic diagram of the proposed spatio-temporal feedback visual system, where each type of specialized neurons in each layer is arranged in matrix form. To simplify the presentation, only one STMD, one LPTC, and their presynaptic neurons are shown. At each time step, an entire image frame is applied to the retina layer and further processed by the four neural layers in a feedforward manner. In the lobula layer, the STMD integrates its output and the output of the LPTC  into the feedback signal to act on medulla outputs via a self-feedback loop.}
	\label{Schematic-Spatio-Temporal-Feedback}
\end{figure*}

\subsection{Retina Layer}
\label{Retina-Layer}
As shown in Fig. \ref{Wiring-Sketch-Spatio-Temporal-Feedback}, the retina layer consists of numerous optical units, ommatidia, each of which views a sector of the whole visual field and sends its axons to the lamina layer. The ommatidia act as luminance receptors with roughly Gaussian sensitivity profile to capture visible light from the natural scene \cite{caves2018visual}. In the proposed visual system, the retina is designed as a two-dimensional array of ommatidia to receive the whole image frame, where each ommatidium is described as a Gaussian filter for yielding a smooth effect on pixel values, as can be seen from Fig. \ref{Schematic-Spatio-Temporal-Feedback}. Mathematically, given an image sequence $I(x,y,t) \in \mathbb{R}$ where $x,y$ is spatial coordinate and $t$ represents time, we formulate the output of an ommatidium $P(x,y,t)$ as
\begin{equation}
	P(x,y,t) =  \iint I(u,v,t) \cdot G_{\sigma_1}(x-u,y-v)dudv
	\label{Photoreceptors-Gaussian-Blur}
\end{equation}
where $G_{\sigma_1}(x,y)$ denotes a Gaussian kernel with standard deviation $\sigma_1$, namely
\begin{equation}
	G_{\sigma_1}(x,y) = \frac{1}{2\pi\sigma_1^2}\exp(-\frac{x^2+y^2}{2\sigma_1^2}).
	\label{Ommatidia-Gaussian-Kernel}
\end{equation}

\subsection{Lamina Layer}
Large monopolar cells (LMCs) receive direct synaptic input from the ommatidia and then carry visual information to the medulla layer, as depicted in Fig. \ref{Wiring-Sketch-Spatio-Temporal-Feedback}. The LMCs are highly sensitive to luminance changes over time, with hyperpolarizing and depolarizing responses to luminance increases and decreases, respectively \cite{tuthill2013contributions}. In the proposed visual system, each LMC is modelled as a temporal filter to calculate luminance changes of each pixel in relation to time (see Fig. \ref{Schematic-Spatio-Temporal-Feedback}). The difference of two Gamma kernels \cite{de1991theory} is adopted as the impulse response of the LMC, considering their excellent temporal processing properties, such as trivial stability, easy adaptation, and the uncoupling of impulse response and filter order. Let $H(t)$ denote the impulse response of the LMC, then we can write it as
\begin{align}
	H(t) &= \Gamma_{n_1,\tau_1}(t) - \Gamma_{n_2,\tau_2}(t) \label{BPF-Impulse-Response}\\
	\Gamma_{n,\tau}(t) &= (nt)^n \frac{\exp(-nt/\tau)}{(n-1)!\cdot \tau^{n+1}} \label{Gamma-Kernel}
\end{align}
where $\Gamma_{n,\tau}(t)$ stands for a Gamma kernel with order $n$ and time constant $\tau$. The output of the LMC $L(x,y,t)$ is defined by convolution of the ommatidium output $P(x,y,t)$ with $H(t)$
\begin{equation}
	L(x,y,t) = \int P(x,y,s) \cdot H(t-s) ds.
	\label{LMCs-BPF}
\end{equation}
Note that luminance of pixel $(x,y)$ will change over time $t$ when an object passes through it. Such temporal change in luminance is reflected in the output of the LMC $L(x,y,t)$. Specifically, the value of the LMC output $L(x,y,t)$ represents the amount of luminance change while the positive and negative signs correspond to luminance increase and decrease, respectively. 



\subsection{Medulla Layer}
Each LMC innervates four medulla neurons, including Tm1, Tm2, Tm3, and Mi1, constituting four parallel information processing pathways \cite{behnia2014processing}, as can be seen from Fig. \ref{Wiring-Sketch-Spatio-Temporal-Feedback}. Specifically, the Mi1 and Tm3 selectively respond to luminance increases with the Mi1 exhibiting a time-delayed response compared with the Tm3; the Tm1 and Tm2 selectively respond to luminance decreases with the Tm1 being temporally-delayed relative to the Tm2. In the proposed visual system, the Tm3 and Tm2 neurons are described as half-wave rectifiers to pass positive part and negative part of the LMC output $L(x,y,t)$, respectively (see Fig. \ref{Schematic-Spatio-Temporal-Feedback}). Denote the output of the Tm3 and Tm2 by $S^{\text{Tm3}}(x,y,t)$ and $S^{\text{Tm2}}(x,y,t)$, respectively, then they can be formulated as
\begin{align}
	S^{\text{Tm3}}(x,y,t) &= [L(x,y,t)]^{+}  \label{Tm3-Output} \\
	S^{\text{Tm2}}(x,y,t) &= [-L(x,y,t)]^{+} \label{Tm2-Output}
\end{align}
where $[x]^+$ represents $\max (x,0)$. The output of the Mi1 and Tm1 neurons denoted by $S_{{(n,\tau)}}^{\text{Mi1}}(x,y,t)$ and $S_{{(n,\tau)}}^{\text{Tm1}}(x,y,t)$, respectively, are formulated as temporally delayed versions of $S^{\text{Tm3}}(x,y,t)$ and $S^{\text{Tm2}}(x,y,t)$, where time delay is achieved by convoluting with a Gamma kernel $\Gamma_{n,\tau}(t)$, as illustrated in Fig. \ref{Schematic-Spatio-Temporal-Feedback}. Formally, $S_{{(n,\tau)}}^{\text{Mi1}}(x,y,t)$ and $S_{{(n,\tau)}}^{\text{Tm1}}(x,y,t)$ are written as  
\begin{align}
	S_{{(n,\tau)}}^{\text{Mi1}}(x,y,t) &= \int [L(x,y,s)]^{+} \cdot \Gamma_{n,\tau}(t-s) ds \label{Mi1-Output}\\
	S_{{(n,\tau)}}^{\text{Tm1}}(x,y,t) &= \int [-L(x,y,s)]^{+} \cdot  \Gamma_{n,\tau}(t-s) ds \label{Tm1-Output}
\end{align}
where time constant $\tau$ and order $n$ of Gamma kernel $\Gamma_{n,\tau}(t)$ control length and order of the time delay unit, respectively.

\subsection{Lobula Layer} 
\label{Lobula-Layer}

As shown in Fig. \ref{Wiring-Sketch-Spatio-Temporal-Feedback}, medulla neural outputs are integrated by two types of lobula neurons, including lobula plate tangential cells (LPTCs) and small target motion detectors (STMDs), which exhibit strong responses to wide-field motion and small target motion, respectively. Furthermore, the response of the STMD is mediated by intra-layer feedback from the LPTC and its self-feedback. In the proposed visual system, the LPTC and STMD are formulated as two motion detectors to extract wide-field motion and small target motion by correlating luminance change patterns from medulla neurons. The intra-layer feedback and the self-feedback loops are designed to embed spatio-temporal background dynamics into the STMD output for eliminating false positive responses in a recurrent manner.

\subsubsection{Small Target Motion Detector}
Each STMD takes input from two medulla neurons located at a single pixel (see Fig. \ref{Schematic-Spatio-Temporal-Feedback}). Specifically, the outputs of the two medulla neurons, i.e., Tm1 and Tm3, are first mediated by subtracting a spatio-temporal feedback signal and then recombined together via a multiplier to generate significant outputs to small target movement, which is written as 
\begin{equation}
	\begin{split}
		D(x,y,t) =&  \Big\{ S^{\text{Tm3}}(x,y,t) -  F(x,y,t) \Big\} \\  & \times 
		\Big\{ S_{{(n_3,\tau_3)}}^{\text{Tm1}}(x,y,t) -  F(x,y,t)\Big\}
		\label{STMD-Signal-Correlation}
	\end{split}
\end{equation}
where $D(x,y,t)$ represents the correlation output of the STMD at pixel $(x,y)$ and time $t$, while $F(x,y,t)$ denotes the spatio-temporal feedback signal formulated as
\begin{equation}
	\begin{split}
		F(x,y,t) = &\alpha \cdot \int_{\Omega} \Big\{D(x-\phi,y-\psi,t-s) \\
		&+ E(x-\phi,y-\psi,t-s) \Big\}\cdot \Gamma_{n_4,\tau_4}(s) ds 
	\end{split}
	\label{Feedback-Signal-Delay} 
\end{equation}
where $\alpha$ is a feedback constant; $\phi(t,s)$ and $\psi(t,s)$ denote the $x$ and $y$ components of background dynamics propagated from the LPTC neurons, respectively, whose formulations are given in the following Sections; $n_4$ and $\tau_4$ are the order and time constant of the Gamma kernel; $E(x,y,t)$ is the weighted summation of the surrounding STMDs' outputs, whose weight function is given by
\begin{equation}
	W_e(x,y)= \frac{1}{2\pi\eta^2}\exp(-\frac{x^2+y^2}{2\eta^2})
	\label{Feedback-Loop-Weight-Function-of-Surrounding-STMDs}
\end{equation}
where constant $\eta$  is controlled by the preferred target sizes of the central STMD, and $E(x,y,t)$ is expressed as
\begin{equation}
	\begin{split}
		E(x,y,t) = \iint S^{\text{Tm3}}&(u,v,t) \cdot S^{\text{Tm1}}(u,v,t) \\
		&\cdot W_e(x-u,y-v) dudv.
	\end{split}
\end{equation}
The correlation output $D(x,y,t)$ is then convolved with a lateral inhibition kernel $W_s(x,y)$ for suppressing responses to large independently-moving objects, that is
\begin{equation}
	Q(x,y,t) = \iint D(u,v,t) \cdot W_s(x-u,y-v) du dv
	\label{STMD-Lateral-Inhibition}
\end{equation}
where $Q(x,y,t)$ represents the STMD output after lateral inhibition and  $W_s(x,y)$ is formulated as
\begin{align}
	W_s(x,y) &= A \cdot [g(x,y)]^{+} + B \cdot [g(x,y)]^{-}  \label{Inhibition-Kernel-W2-p1}\\
	g(x,y)  &= G_{\sigma_2}(x,y) - e \cdot G_{\sigma_3}(x,y) - \rho
	\label{Inhibition-Kernel-W2-p2}
\end{align}
where $[x]^+$ and $[x]^-$ represent $\max (x,0)$ and $\min (x,0)$, respectively, $A$, $B$, $e$ and $\rho$ are constant.

\begin{figure}[t!]
	\centering
    \hspace{3.5mm}
	\subfloat[]{\includegraphics[width=0.16\textwidth]{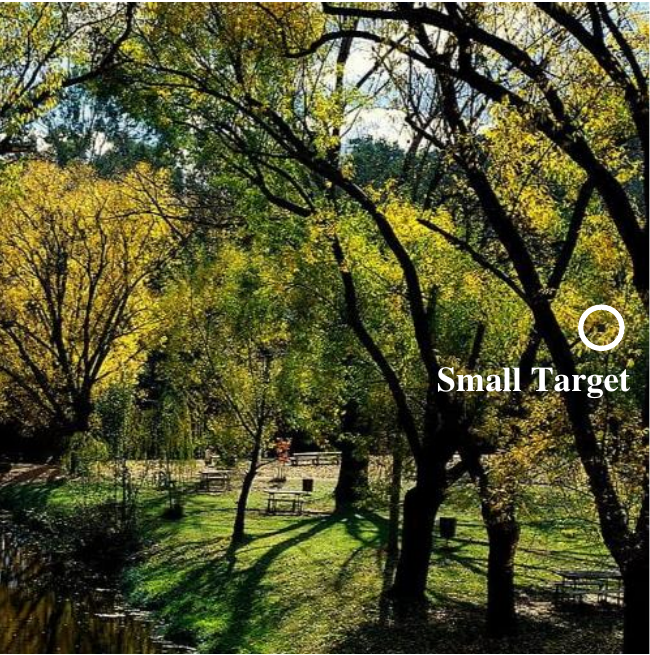}}
	\hfil
	\subfloat[]{\includegraphics[width=0.28\textwidth]{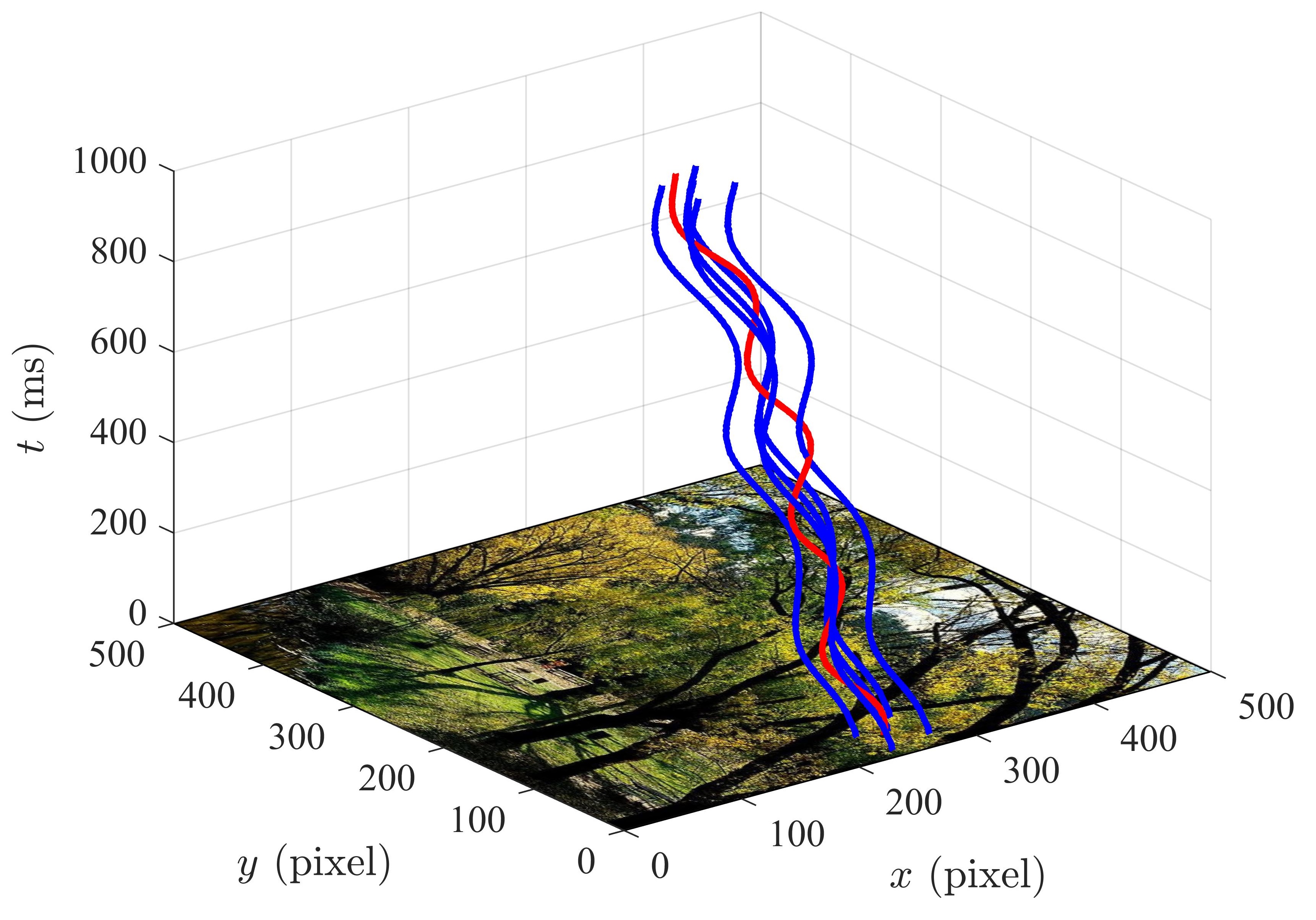}}
	\hfil
	\subfloat[]{\includegraphics[width=0.235\textwidth]{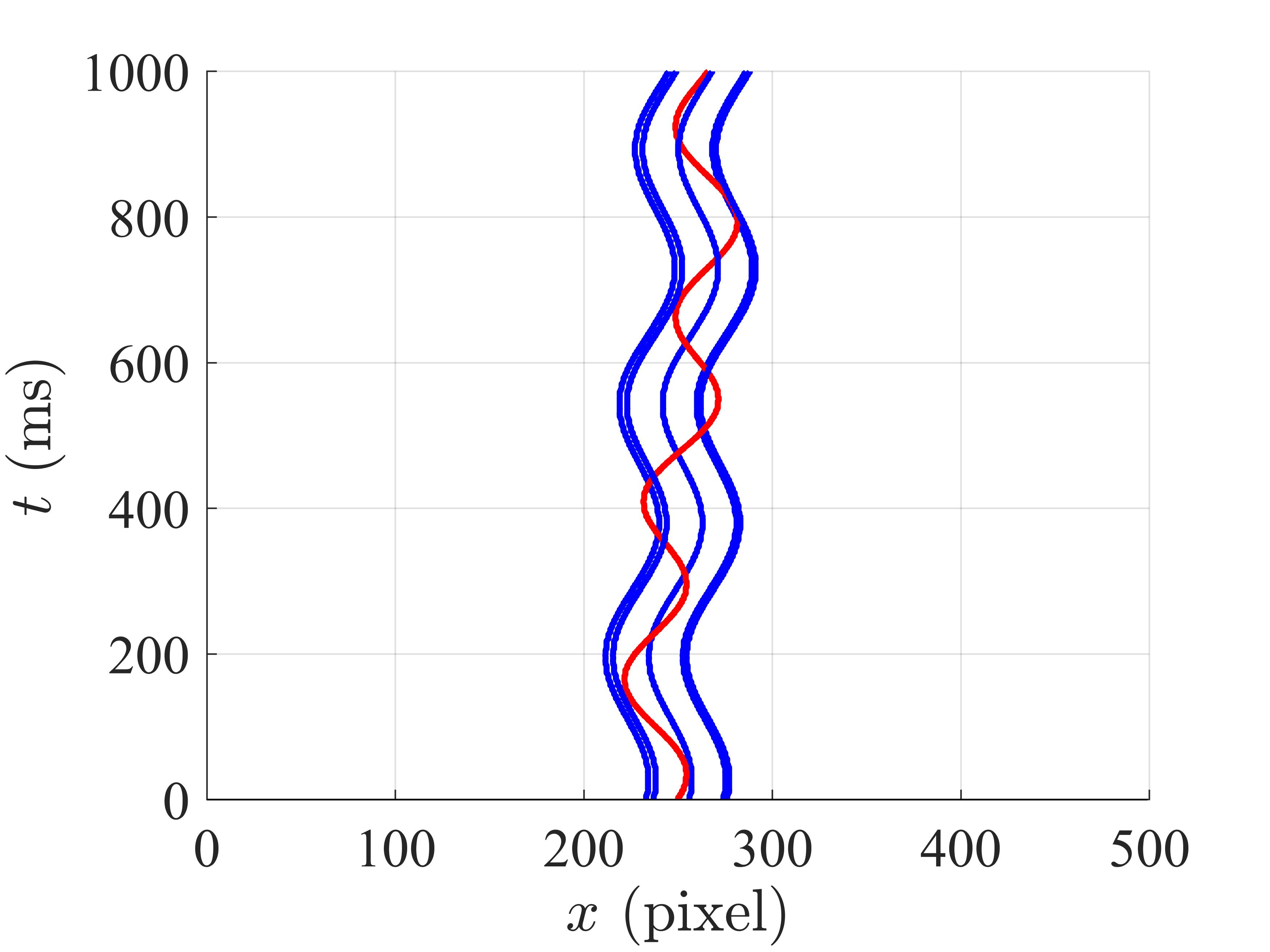}}
	\hfil
	\subfloat[]{\includegraphics[width=0.235\textwidth]{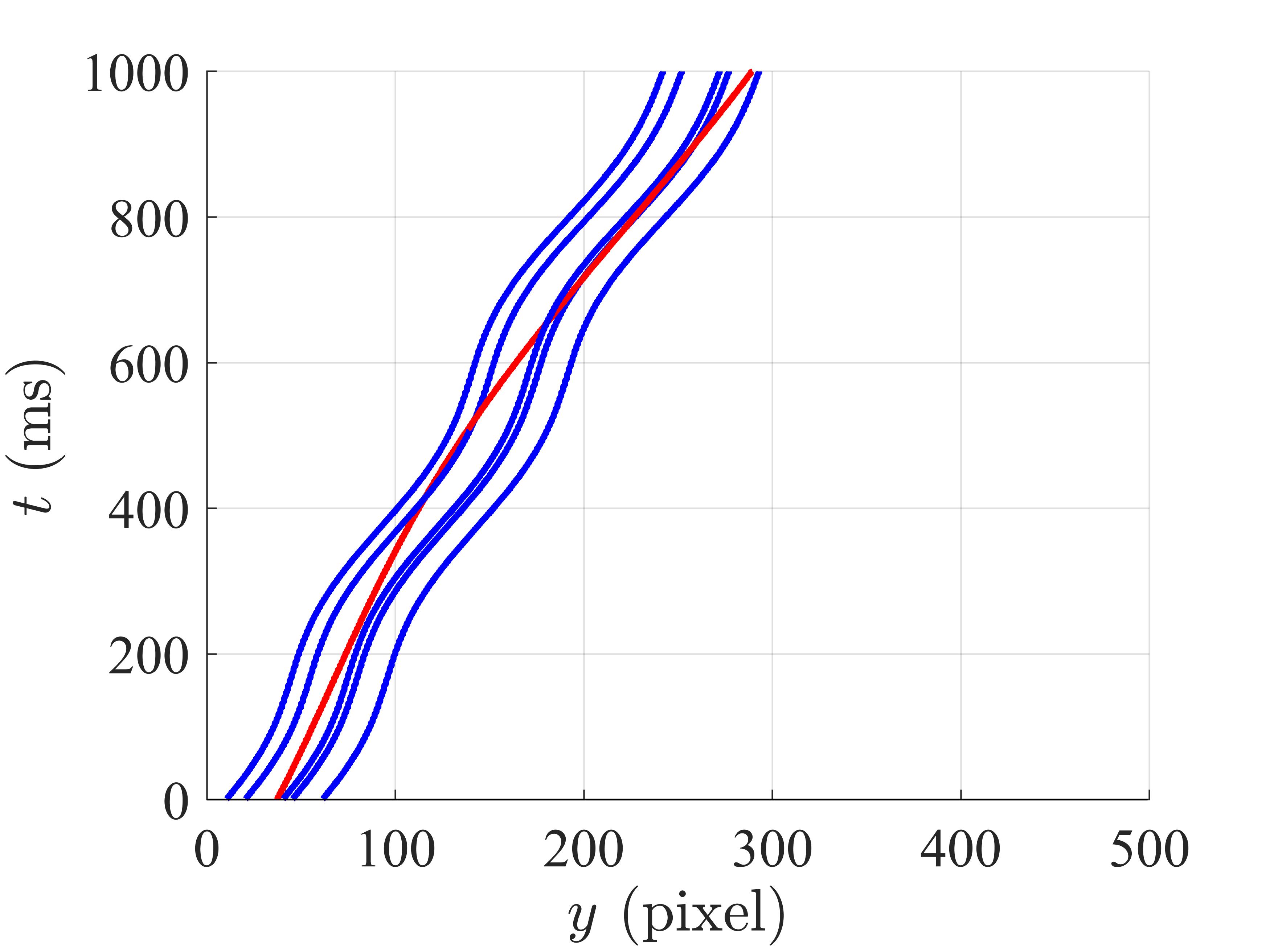}}
	\caption{Example of (a) an image frame and object trajectories observed in (b) spatio-temporal space, (c) $x$-$t$ plane, and (d) $y$-$t$ plane, respectively. For the clarity of presentation, only one small target trajectory (red)  and a few background object trajectories (blue) are shown. The red trajectory (small target) can be clearly discriminated from the blue trajectories (background objects) in terms of $x$ and $y$ components over time $t$, even though they seem to be extremely cluttered in spatio-temporal space. }
	\label{Schematic-Background-Target-Point-Trajectory-3D-2D}
\end{figure} 


\subsubsection{Motivation of Spatio-Temporal Feedback}
\label{Motivation-of-Spatio-Temporal-Feedback} 
Motion is generally more homogeneous within an object region, compared to other visual features, such as color, texture, and luminance \cite{keuper2018motion,lezama2011track}. Grouping motion clusters that share similar spatio-temporal characteristics in an image sequence would provide a powerful cue for object discrimination \cite{gallego2018unifying,shen2018submodular, luria2014come}. Moreover, point trajectories spanning multiple frames are always more robust to short-term variations compared to two-frame motion fields in the task of motion segmentation \cite{gollisch2010eye,shi2013robust}. Fig. \ref{Schematic-Background-Target-Point-Trajectory-3D-2D} illustrates trajectories of a small target and its surrounding background in spatio-temporal space. The small target has relative movement to the complex background, so its trajectory is dissimilar to those of surrounding background, which could be clearly discriminated by observing $x$ and $y$ components of trajectories with respect to time $t$, as shown in Fig. \ref{Schematic-Background-Target-Point-Trajectory-3D-2D}(b) and (c). Motivated by this, we design $\phi(t,s), \psi(t,s)$ to reflect spatial and temporal dynamics of background motion, which correspond to $x$ and $y$ components of background trajectories, respectively. Substituting $\phi(t,s)$ and $\psi(t,s)$ into the spatio-temporal feedback signal $F(x,y,t)$ given by (\ref{Feedback-Signal-Delay}), the STMD responses to false positives with similar motion dynamics to complex background would be largely suppressed after the negative feedback. In the following section, lobula plate tangential cell (LPTC) model and its population coding mechanism are proposed to obtain $\phi(t,s)$ and $\psi(t,s)$. 

\subsubsection{Lobula Plate Tangential Cell}
\label{Lobula-Plate-Tangential-Cell} 
\begin{figure}[t!]
	\centering
	\subfloat[]{\includegraphics[width=0.4\textwidth]{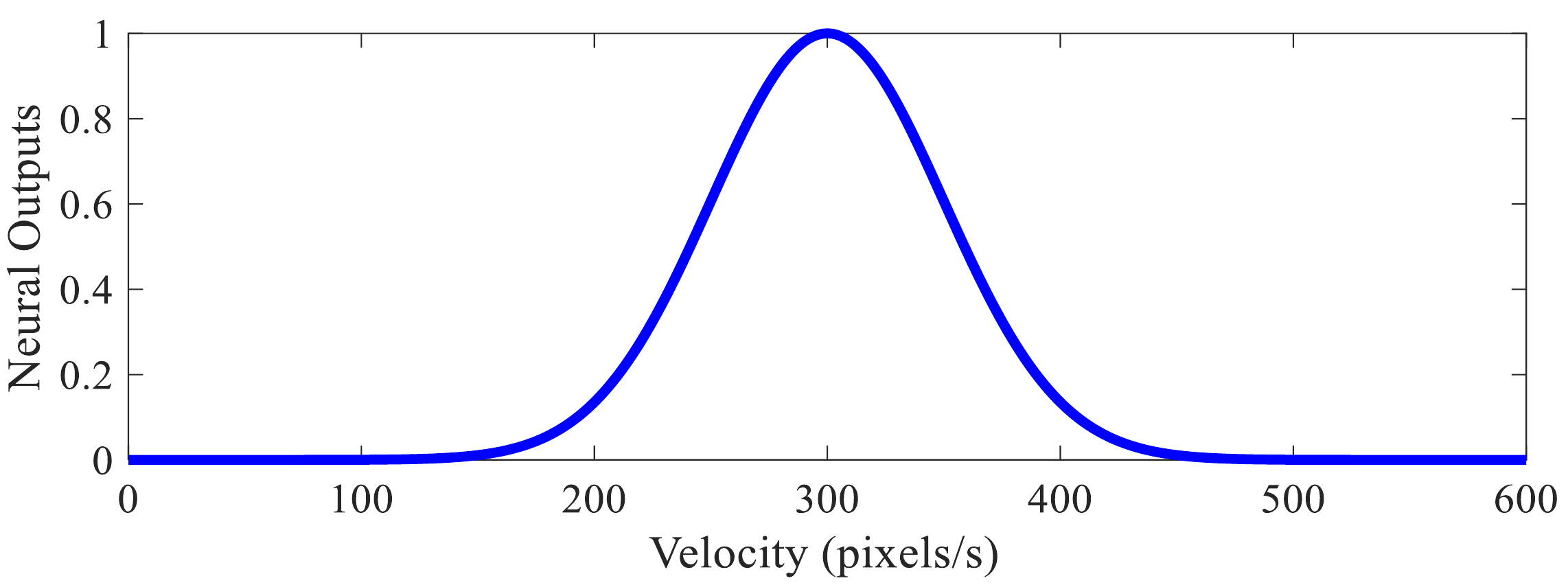}}
	\hfil
	\subfloat[]{\includegraphics[width=0.4\textwidth]{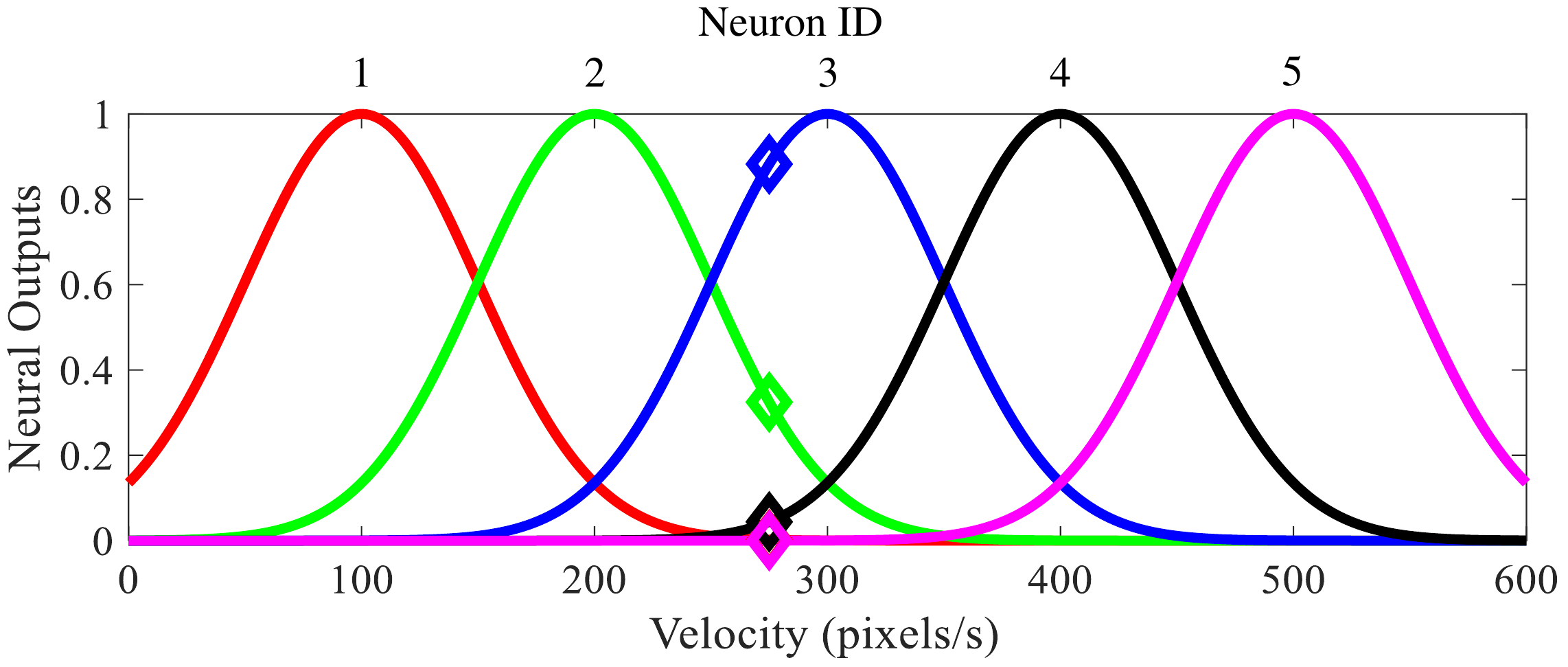}}
	\hfil
	\subfloat[]{\includegraphics[width=0.385\textwidth]{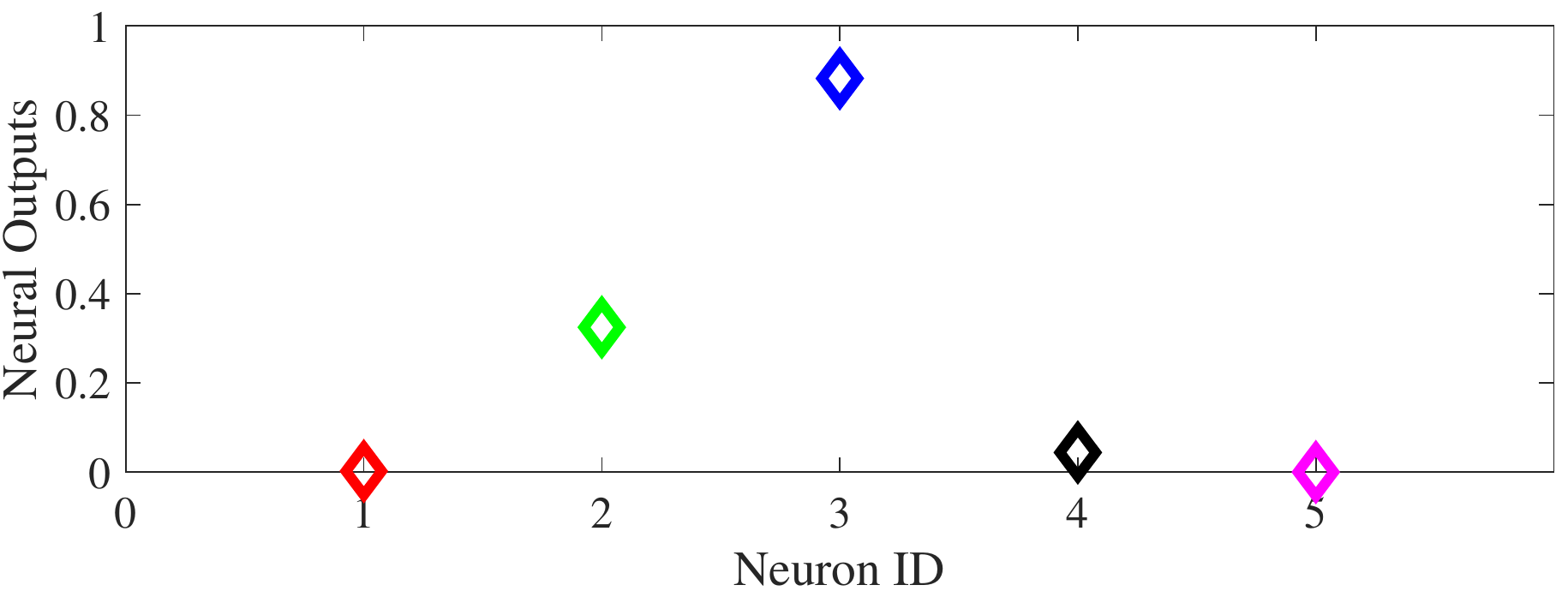}}
	\caption{(a) Tuning curve of a single neuron with respect to object velocity. The neuron responds to objects with velocities in a specific range roughly $[150,450]$ pixels/s, and its output peaks at an optimal velocity $300$ pixels/s. (b) Schematic of neural population coding based on five neurons with different preferred velocity ranges. For an object with velocity $275$ pixels/s, the intersection points with each tuning curve are arranged into a vector to yield the representation of object velocity as $\{r_k |k=1,2,\cdots,5\} = \{0.0, 0.32, 0.88, 0.04, 0.0\}$. (c) Firing rate vector of all velocity tuned neurons $\{r_k |k=1,2,\cdots,5\}$.
	}
	\label{LPTC-Neuron-Tuning-Curve-Population-Coding}
\end{figure}

Each LPTC collects outputs of medulla neurons located at two different pixels for wide-field motion detection, as shown in Fig. \ref{Schematic-Spatio-Temporal-Feedback}. Denote the two pixels by $(x,y)$ and $(x'(\beta, \theta),y'(\beta, \theta))$, respectively, then they are written as
\begin{equation}
	\begin{split}
		x'(\beta, \theta) &= x + \beta \cdot \cos\theta \\ 
		y'(\beta, \theta) &= y + \beta \cdot \sin\theta
	\end{split}
	\label{Formulation-Two-Pixels-X-Y}
\end{equation}
where $\beta$ is the distance between the two pixels and $\theta$ denotes the preferred direction of the LPTC. Multiplication of luminance-increase signals at these two pixels is summed together with that of luminance-decrease signals to define the output of the LPTC, formulated as
\begin{equation}
	\begin{split}
		R(x,y,t,\beta,\theta) = & S^{\text{Tm3}}(x,y,t) \cdot S^{\text{Mi1}}_{{(n_{_5},\tau_{_5})}}(x',y',t) \\  
		& + S^{\text{Tm2}}(x,y,t) \cdot S^{\text{Tm1}}_{{(n_{_5},\tau_{_5})}}(x',y',t)
	\end{split}
	\label{LPTC-Neuron-Correlation}
\end{equation}
where $R(x,y,t,\beta,\theta)$ stands for the output of the LPTC; the time constants $\tau_5$ is determined by time interval between the luminance-increase signals (or luminance-decrease signals) of the two pixels; the order $n_5$ controls the shapes of signals after the time delay. 

Each LPTC is highly selective to object velocity. Specifically, it exhibits responses to the objects with velocities in a specific range and reaches the strongest output at an optimal velocity, as illustrated in Fig. \ref{LPTC-Neuron-Tuning-Curve-Population-Coding}(a). In animals' visual systems, object velocity is believed to be encoded by neural ensembles rather than an individual neuron \cite{koay2022sequential,glaser2018population}. Taking inspiration from the neural population coding mechanism, we design a collection of the LPTC neurons with overlapping preferred velocity ranges to encode object velocity into neural activities, as shown in Fig. \ref{LPTC-Neuron-Tuning-Curve-Population-Coding}(b) and (c). More precisely, the output of the LPTC selective to a specific velocity range is obtained by adjusting the correlation distance between the two pixels, i.e., $\beta$ in (\ref{LPTC-Neuron-Correlation}), whose sensitivity analysis is discussed in Section \ref{Parameter-Sensitivity-Study}. We properly select a set of the correlation distance denoted by $\{\beta_i | i = 1,2,\cdots, N\}$ to ensure the preferred velocity ranges of the LPTCs cover the object velocity range. Then the firing rate of the $i$th LPTC in response to a moving object is defined as the strongest 
output regarding to direction $\theta$, that is
\begin{align}
	r(t,\beta_i) = \iint_{\Omega} R(x,y,t,\beta_i,\tilde{\theta}) dx dy \\
	\tilde{\theta} = \arg \max_{\theta} \iint_{\Omega} R(x,y,t,\beta_i,\theta) dx dy
\end{align}
where $r(t,\beta_i)$ denotes the firing rate of the $i$th LPTC and $\tilde{\theta}$ represents the direction of the strongest output. The firing rate vector $\{r(t,\beta_i) | i = 1,2,\cdots, N \}$ is further compared with the tuning curves of the LPTCs to estimate the velocity of a moving object, that is
\begin{align}
	\nu(t) = \arg \max_{v} \prod_{i} \exp \big{(}-||r(t,\beta_i) -  f(v,\beta_i)||^2  \big{)}
\end{align} 
where $\nu(t)$ denotes the velocity of the moving object at time $t$ and $f(v,\beta_i)$ represents the tuning curve of the $i$th LPTC. The $x$ and $y$ components of the object trajectory, i.e., $\phi(t,s)$ and $\psi(t,s)$, are defined using the accumulation of  $\nu(t)$ during time period $[s, t]$, that is 
\begin{align}
	\phi(t,s) & = \int_s^t \nu(\tau) \cos\tilde{\theta} d\tau \\
	\psi(t,s) & = \int_s^t \nu(\tau) \sin\tilde{\theta} d\tau.
\end{align}
The obtained $\phi(t,s)$ and $\psi(t,s)$ is further fed into the STMD subnetwork to incorporate in the spatio-temporal feedback signal $F(x,y,t)$.

\section{Experiments}
\label{Experiments}

\subsection{Experimental Setup}

\subsubsection{Datasets}
We quantify the performance of the developed spatio-temporal feedback visual system on Vision Egg dataset \cite{straw2008vision} and RIST dataset  \cite{RIST-Data-Set} for small target motion detection. Vision Egg is a large collection of video clips each of which holds one or multiple synthetic small targets moving against complex natural backgrounds. It contains a variety of background scene images and small targets categorized by luminance, velocity, and size. The spatial resolution of the videos range from $200 \times 200$ to $500 \times 500$ pixels, while their temporal resolutions are equal to $1000$ fps. RIST is a challenging dataset consisting of $21$ real-world video clips recorded by GoPro action camera with a spatial resolution of $480 \times 270$ pixels at $240$ fps. Its scenarios involve many types of challenges, such as non-uniform dynamic shadows, heavy background clutters, sudden camera movement, various weather conditions, and illumination variations. The size of small targets in the captured videos ranges from $3\times 3$ to $15 \times 15$ pixels.

\subsubsection{Evaluation Metrics}
We take detection rate ($D_R$) and false alarm rate ($F_A$) as metrics to measure the detection performance. Higher detection rate and lower false alarm rate denote better detection performance. The metrics can be formulated as follow
\begin{equation}
	D_R = \frac{N_t}{N_a}, F_A = \frac{N_f}{N_F}
\end{equation}
where $N_t$ is the number of true positives, $N_a$ is the number of actual targets, $N_f$ is the number of false positives, and $N_F$ is the number of frames. A detected position is declared as a true positive if its distance to a ground truth is within $5$ pixels. To assess detection performance under different detection thresholds, we adopt a receiver operating characteristic curve (ROC) which plots detection rate ($D_R$) against false positive rate ($F_A$) with a varying detection threshold.

\subsubsection{Implementation Details}
\begin{table}[t!]
	\renewcommand{\arraystretch}{1.3}
	\caption{Parameters of the Spatio-Temporal Feedback Visual System}
	\label{Table-Parameter-Spatio-Temporal-Feedback-Neural-Network}
	\centering
	\begin{tabular}{cc}
		\hline
		Eq. & Parameters \\	
		\hline
		(\ref{Photoreceptors-Gaussian-Blur}) & $\sigma_1 = 1$ \\
		
		(\ref{BPF-Impulse-Response}) & $n_1 = 2, \tau_1= 3, n_2 = 6, \tau_2 = 9$\\
		
		(\ref{STMD-Signal-Correlation}) & $n_3 = 5, \tau_3 = 25$ \\
		
		(\ref{Feedback-Signal-Delay}) & $\alpha = 0.1, n_4 = 6, \tau_4 = 12$ \\
		
		(\ref{Feedback-Loop-Weight-Function-of-Surrounding-STMDs}) & $\eta = 1.5$ \\
		
		(\ref{Inhibition-Kernel-W2-p1}) & $A = 1, B = 3$ \\
		
		(\ref{Inhibition-Kernel-W2-p2}) & $\ \sigma_2 = 1.5, \sigma_3 = 3, e = 1, \rho = 0$ \\
		
		(\ref{Formulation-Two-Pixels-X-Y}) & $\beta \in \{2, 4, 6, 8, 10, 12, 14, 16, 18\} $ \\
		
		(\ref{LPTC-Neuron-Correlation}) & $n_5 = 25, \tau_5 = 30$ \\
		\hline
	\end{tabular}
\end{table}	
Parameters of the developed spatio-temporal feedback visual system are shown in Table \ref{Table-Parameter-Spatio-Temporal-Feedback-Neural-Network}, which can be roughly classified into two groups for the STMD and LPTC subnetworks, respectively. The sensitivity analysis for the parameters of the STMD subnetwork and their effect on detection performance have been investigated in the previous works \cite{wang2021time,wang2018directionally}. In the experiments, we properly tune these parameters to endow the STMD subnetwork with similar response properties to the STMD neurons. As for the LPTC subnetwork, its performance is controlled by three parameters, including the correlation distance $\beta$, order $n_5$, and time constant $\tau_5$ [see (\ref{Formulation-Two-Pixels-X-Y}) and (\ref{LPTC-Neuron-Correlation})]. Tuning these parameters will change the preferred velocity range of the LPTC. We further conduct a sensitivity study of these parameters in Section \ref{Parameter-Sensitivity-Study} to evaluate their effect on preferred velocity range and provide a guidance for parameter setting of the population coding mechanism.


\begin{figure*}[!t]
	\vspace{-7.0mm}
	\centering
	\subfloat[]{\includegraphics[width=0.24\textwidth]{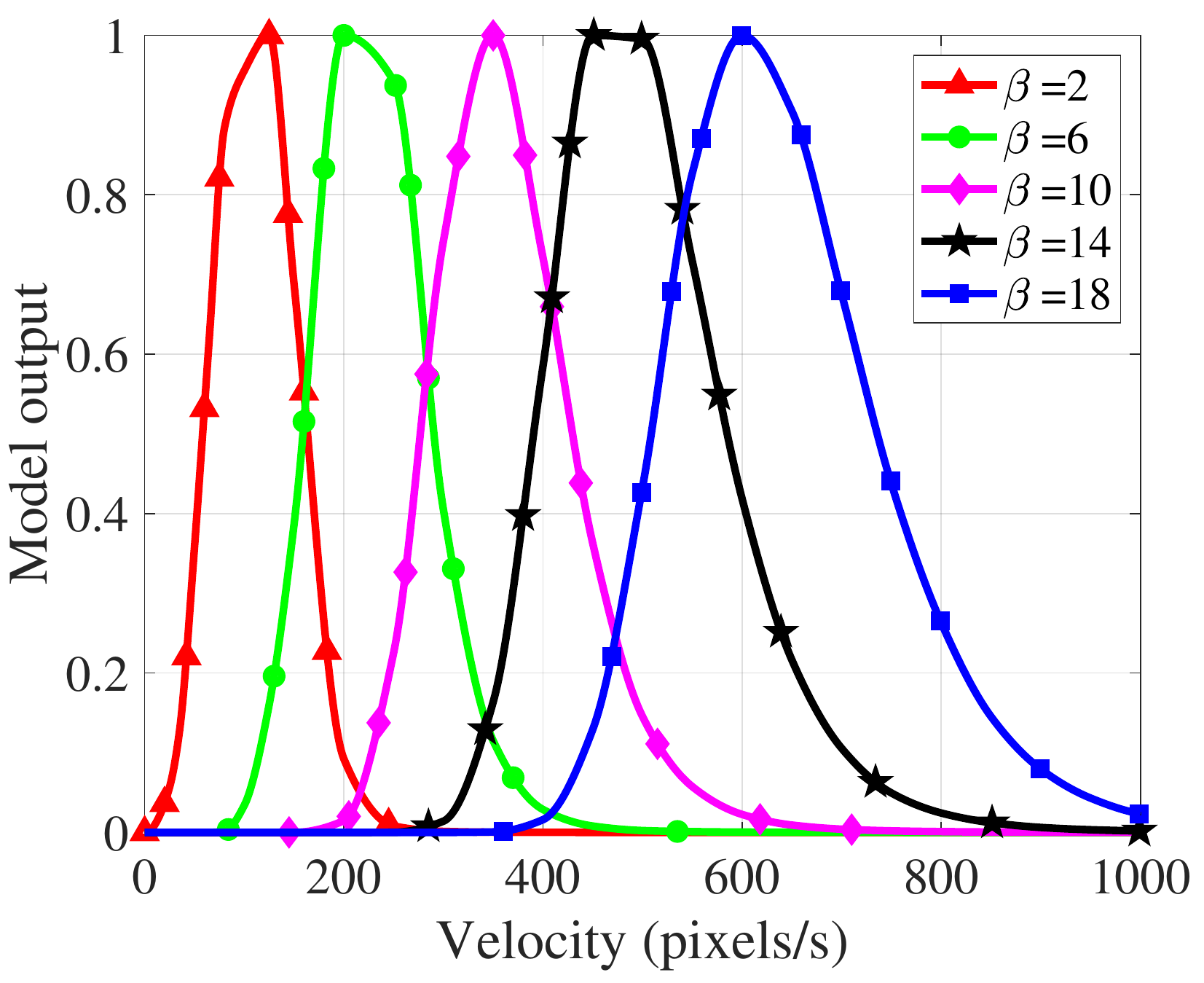}}
	\hfil
	\subfloat[]{\includegraphics[width=0.24\textwidth]{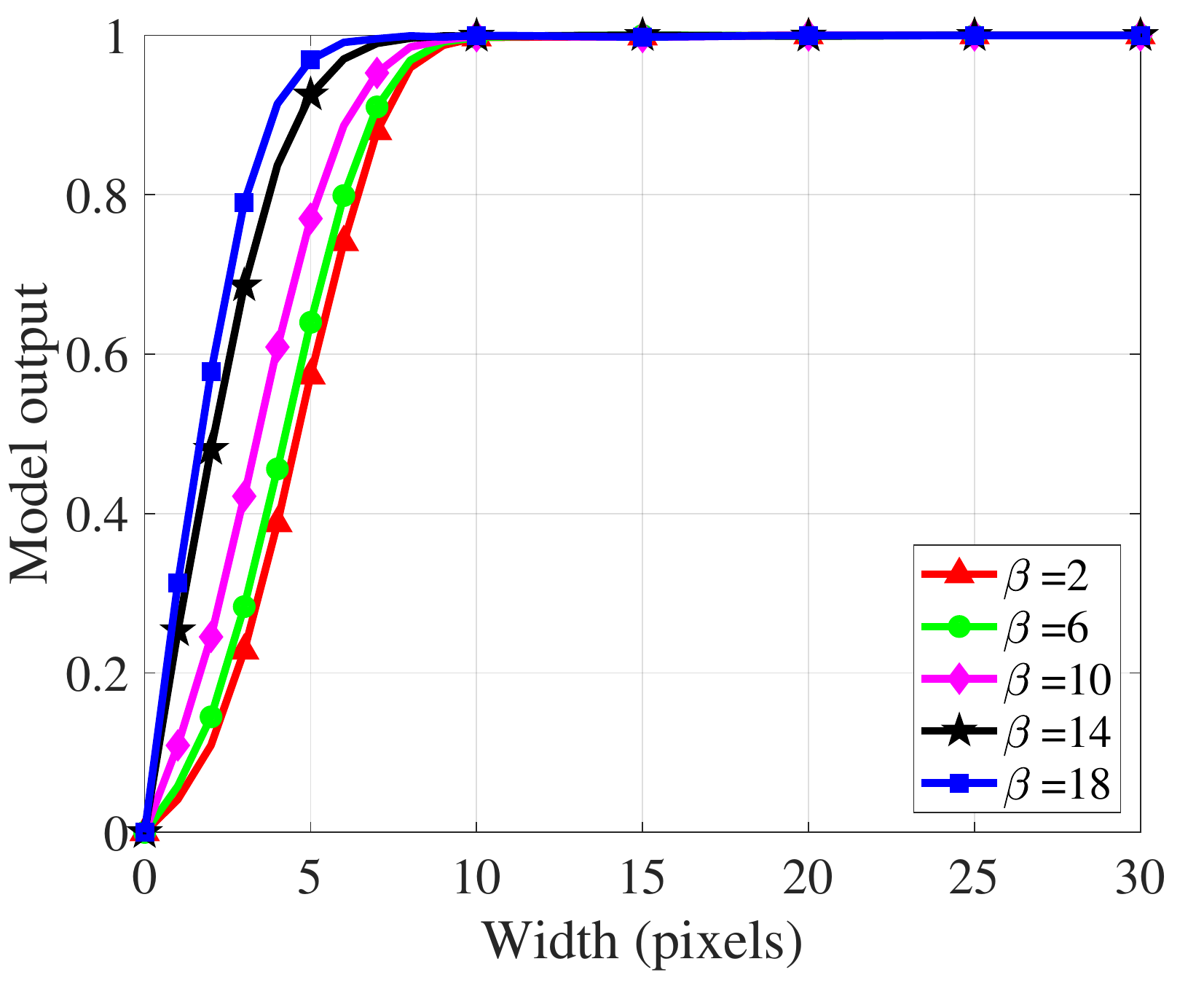}}
	\hfil
	\subfloat[]{\includegraphics[width=0.24\textwidth]{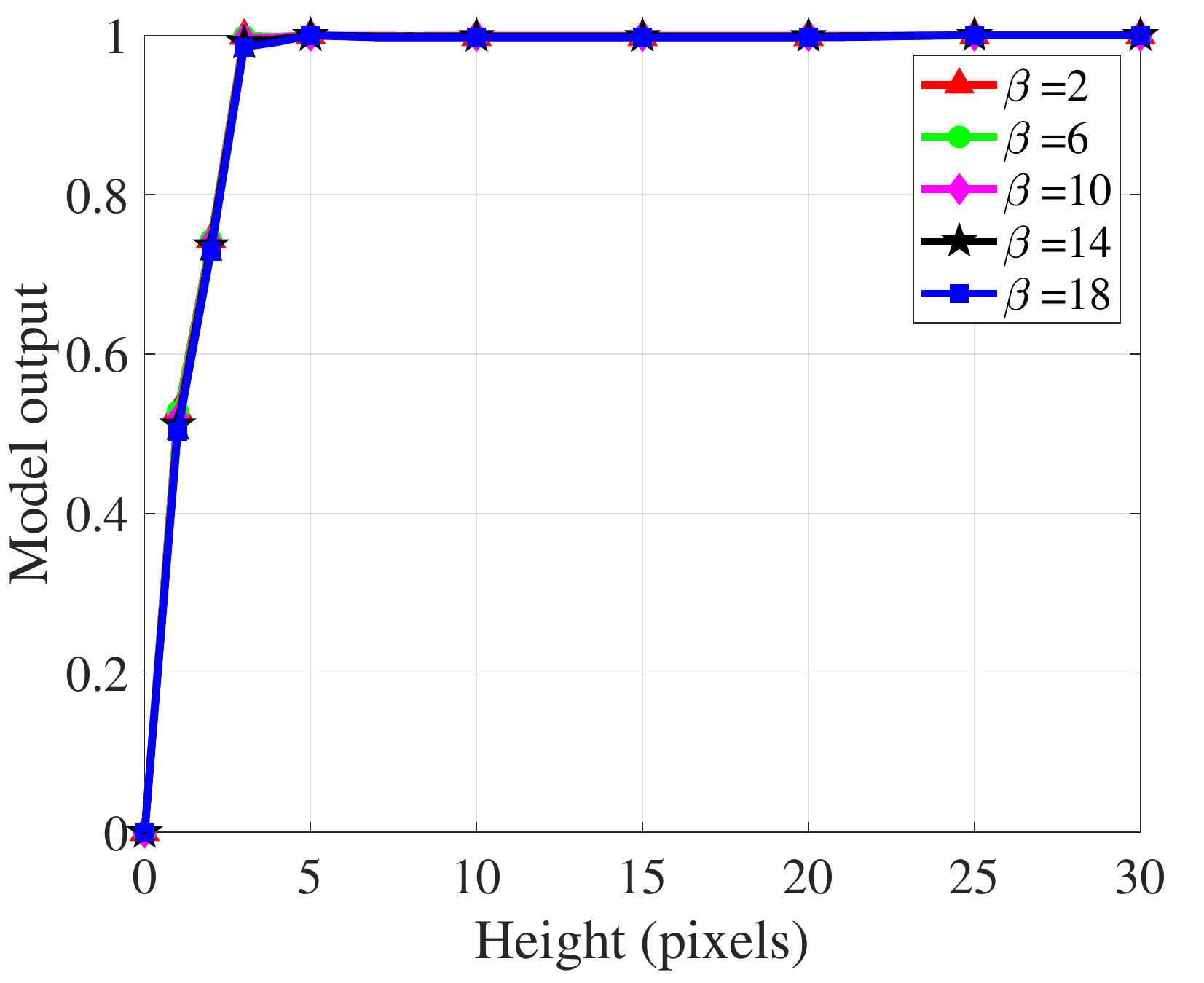}}
	\hfil
	\subfloat[]{\includegraphics[width=0.24\textwidth]{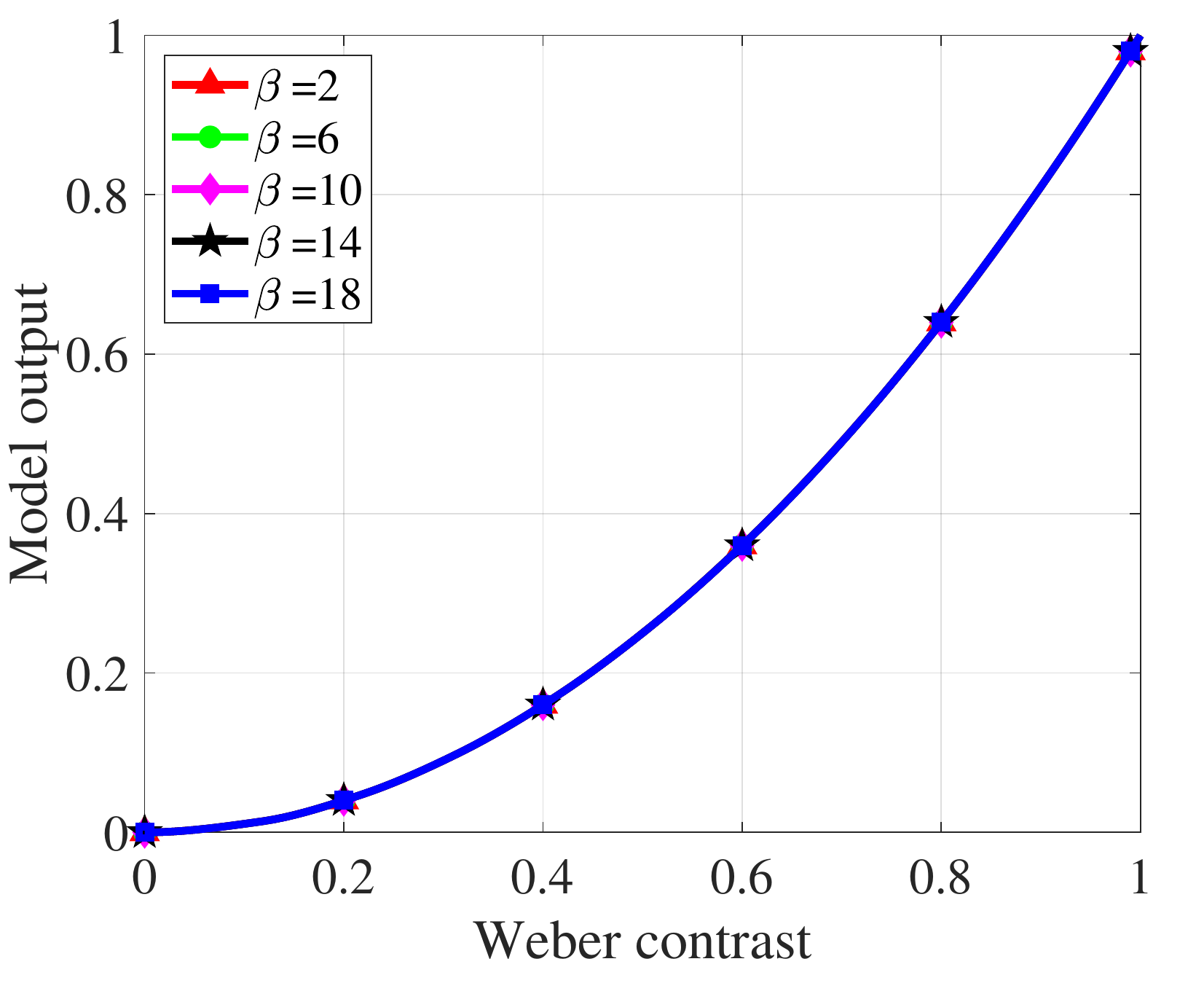}}
	\caption{Outputs of the LPTC neural model at various setting of the correlation distance $\beta$ regarding to (a) velocity, (b) width, (c) height, (d) Weber contrast of the object.}
	\label{Five-EMDs-Tuning-Curve-Marker}
\end{figure*}

\begin{figure*}[!t]
	\vspace{-4.0mm}
	\centering
	\subfloat[]{\includegraphics[width=0.24\textwidth]{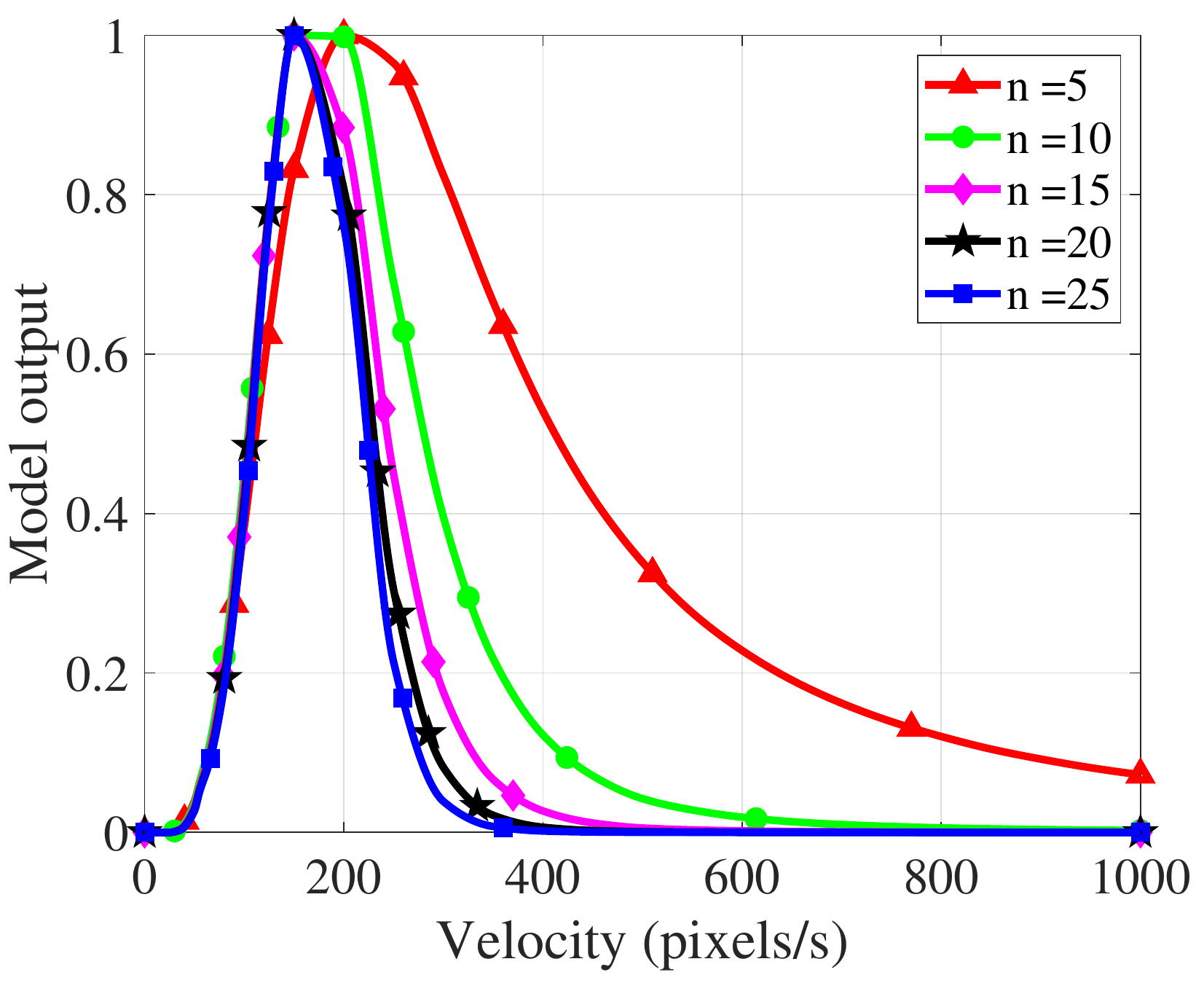}}
    \hfil
     \subfloat[]{\includegraphics[width=0.24\textwidth]{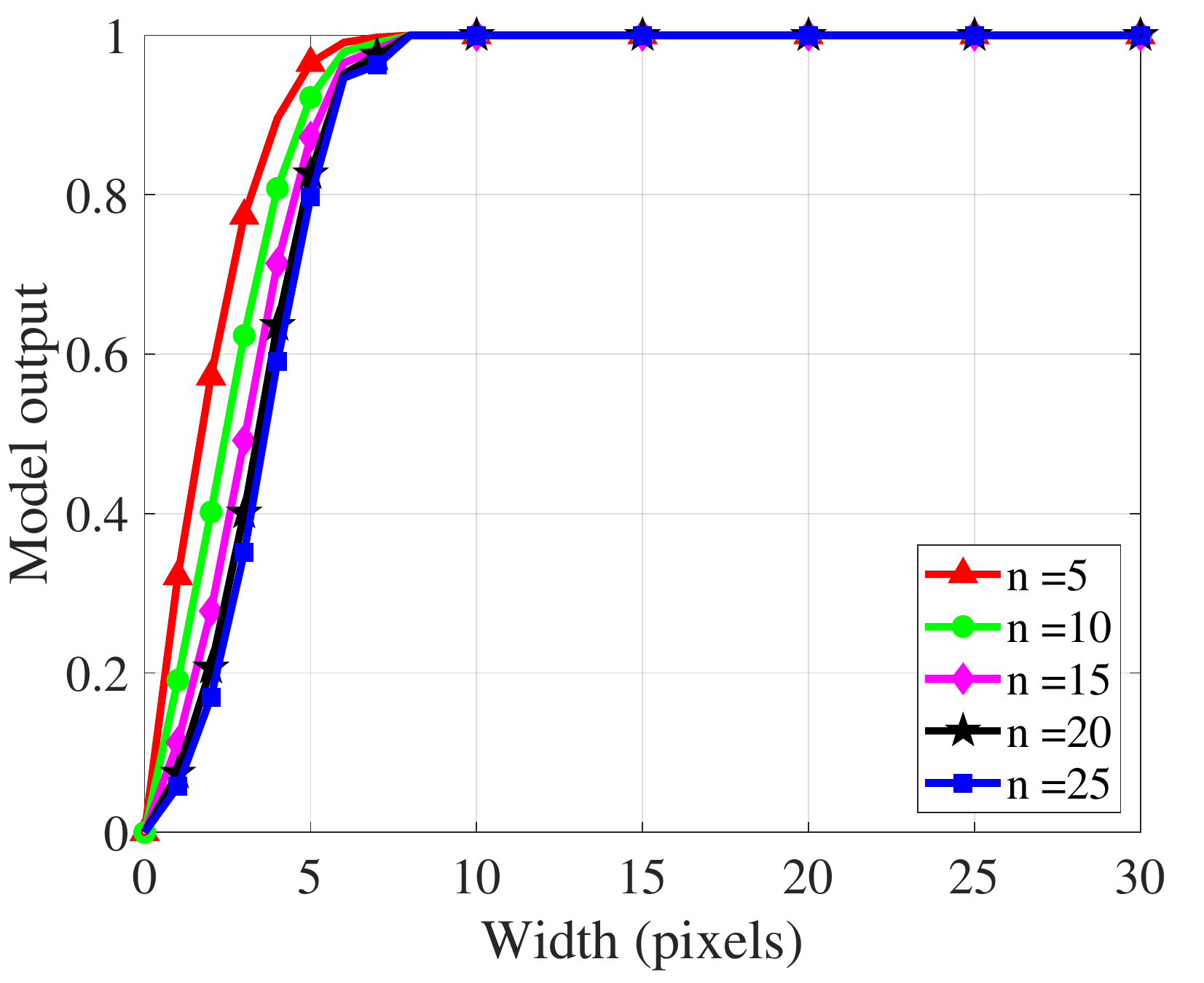}}
    \hfil
	\subfloat[]{\includegraphics[width=0.24\textwidth]{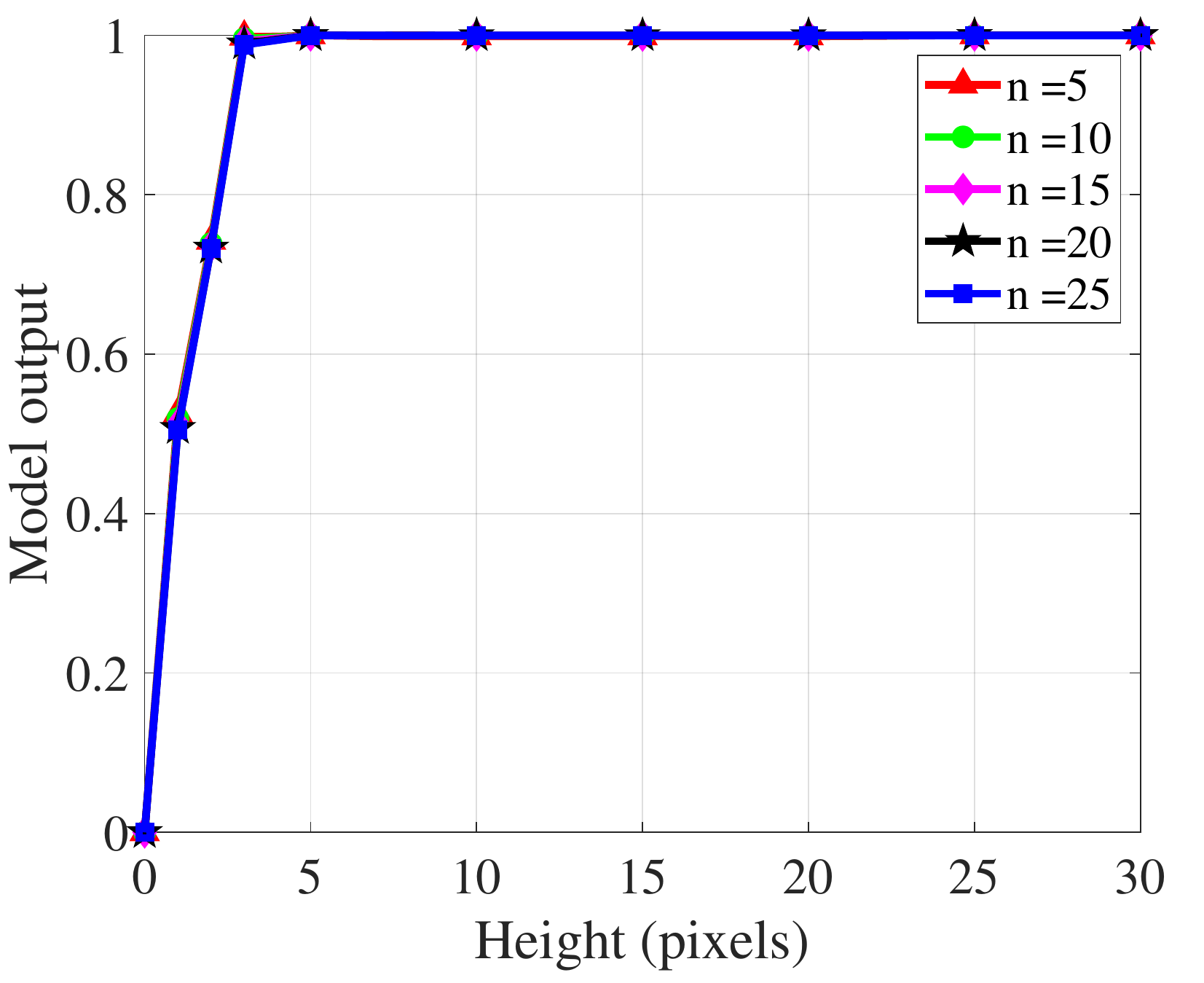}}
	\hfil
	\subfloat[]{\includegraphics[width=0.24\textwidth]{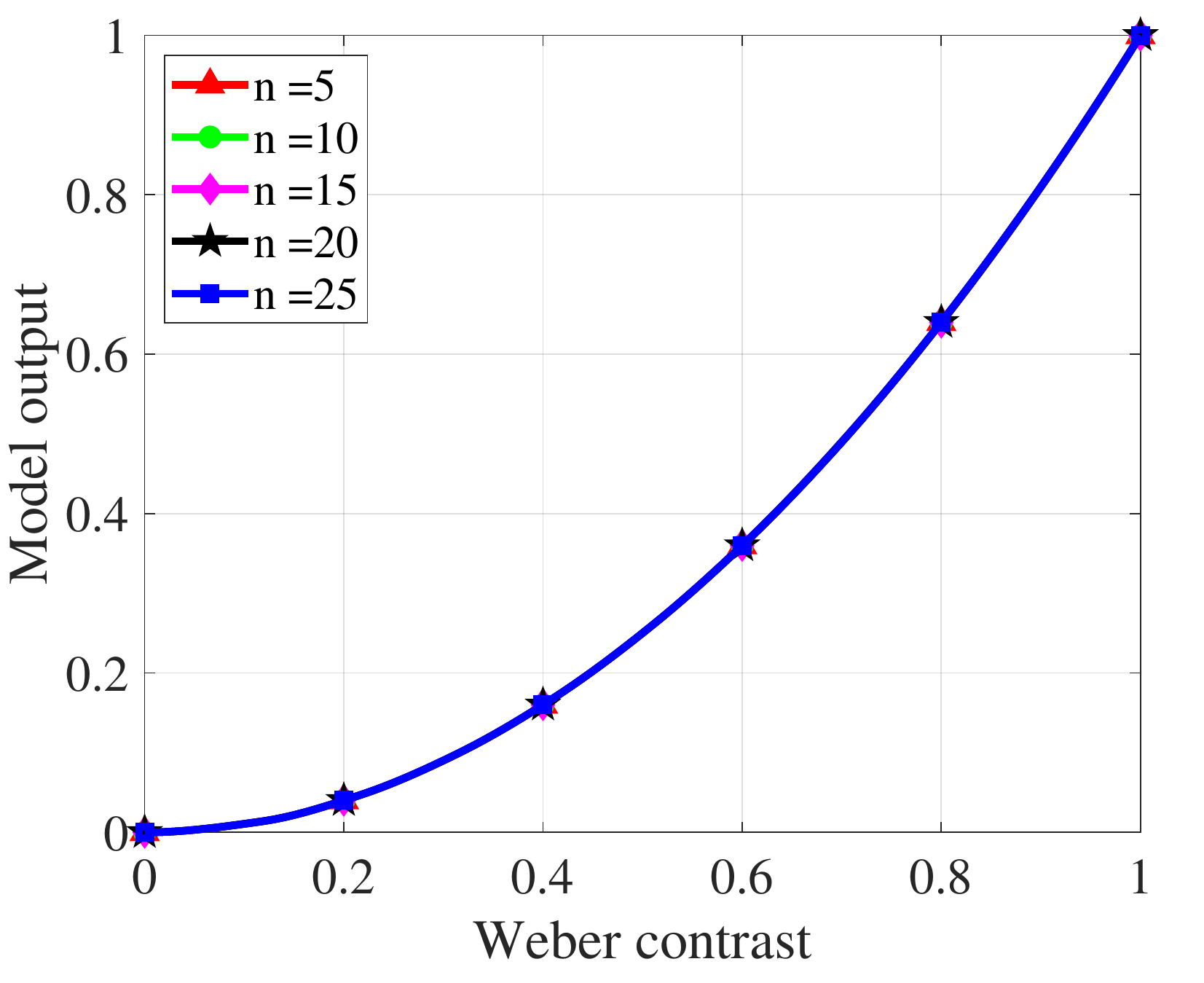}}
	\caption{Outputs of the LPTC neural model at various setting of the order $n$ regarding to (a) velocity, (b) width, (c) height, (d) Weber contrast of the object.}
	\label{EMD-Varying-Order-Tau-30-Marker}
\end{figure*}

\begin{figure*}[!t]
	\vspace{-4.0mm}
	\centering
	\subfloat[]{\includegraphics[width=0.24\textwidth]{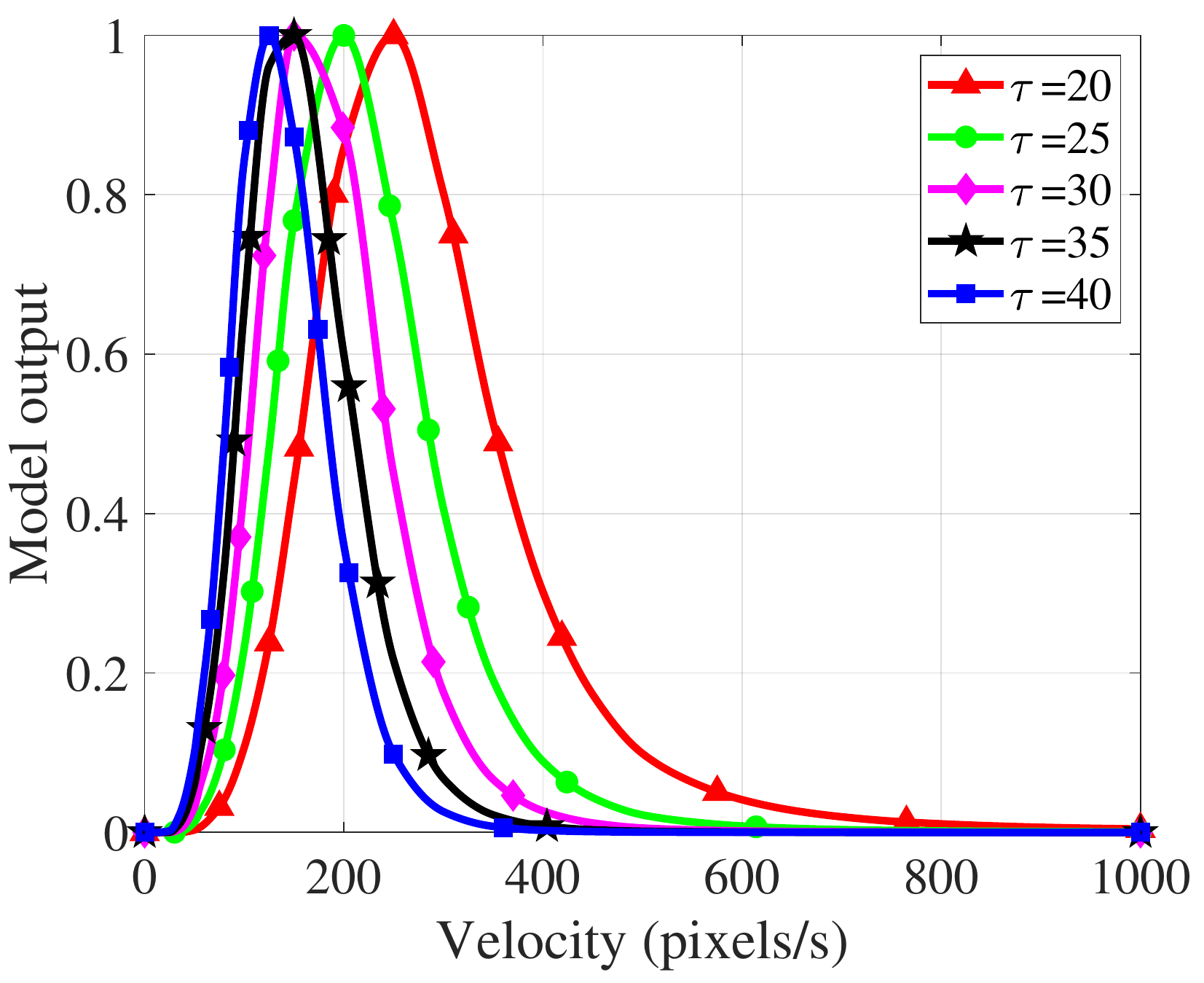}}
	\hfil
	\subfloat[]{\includegraphics[width=0.24\textwidth]{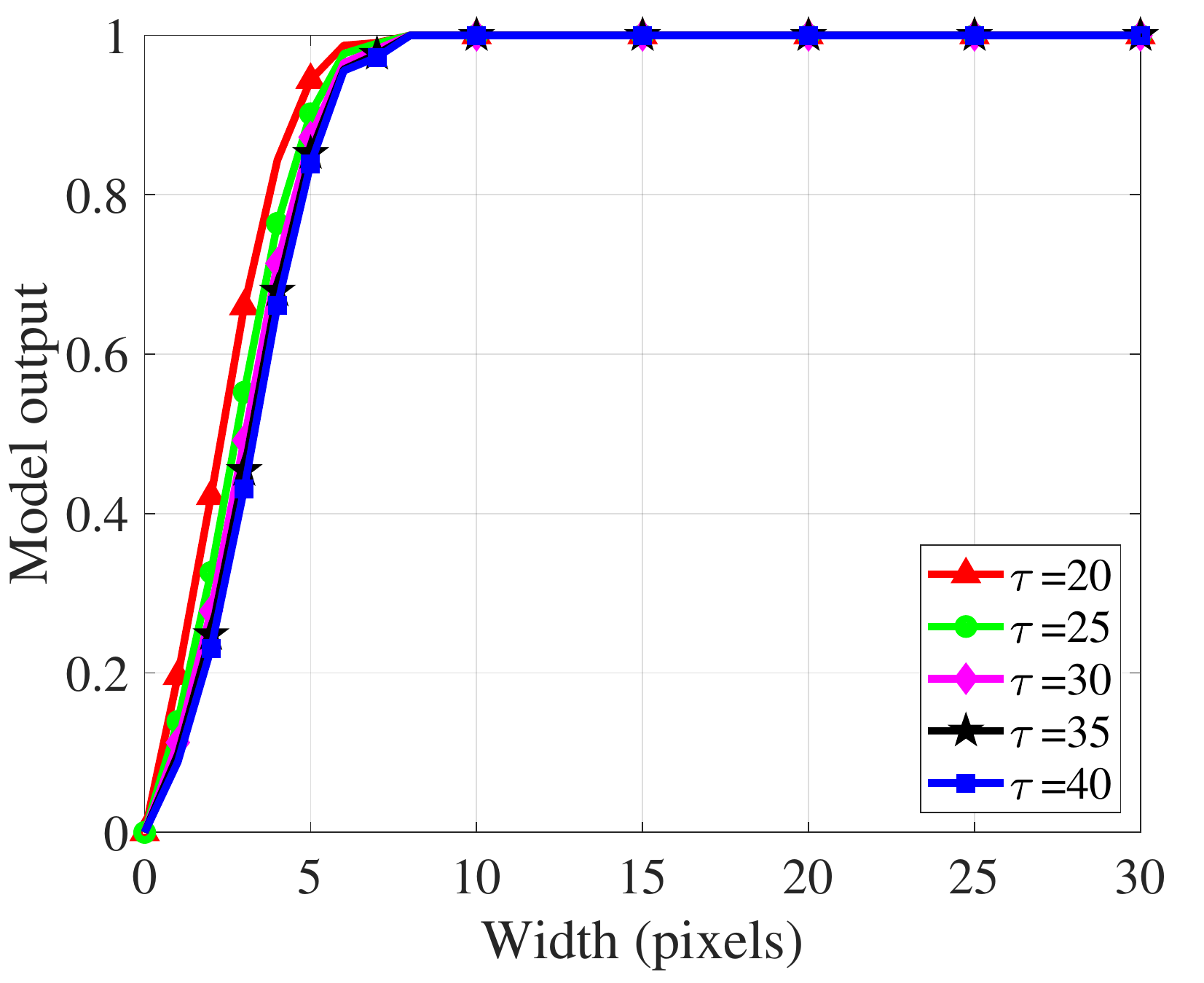}}
	\hfil
	\subfloat[]{\includegraphics[width=0.24\textwidth]{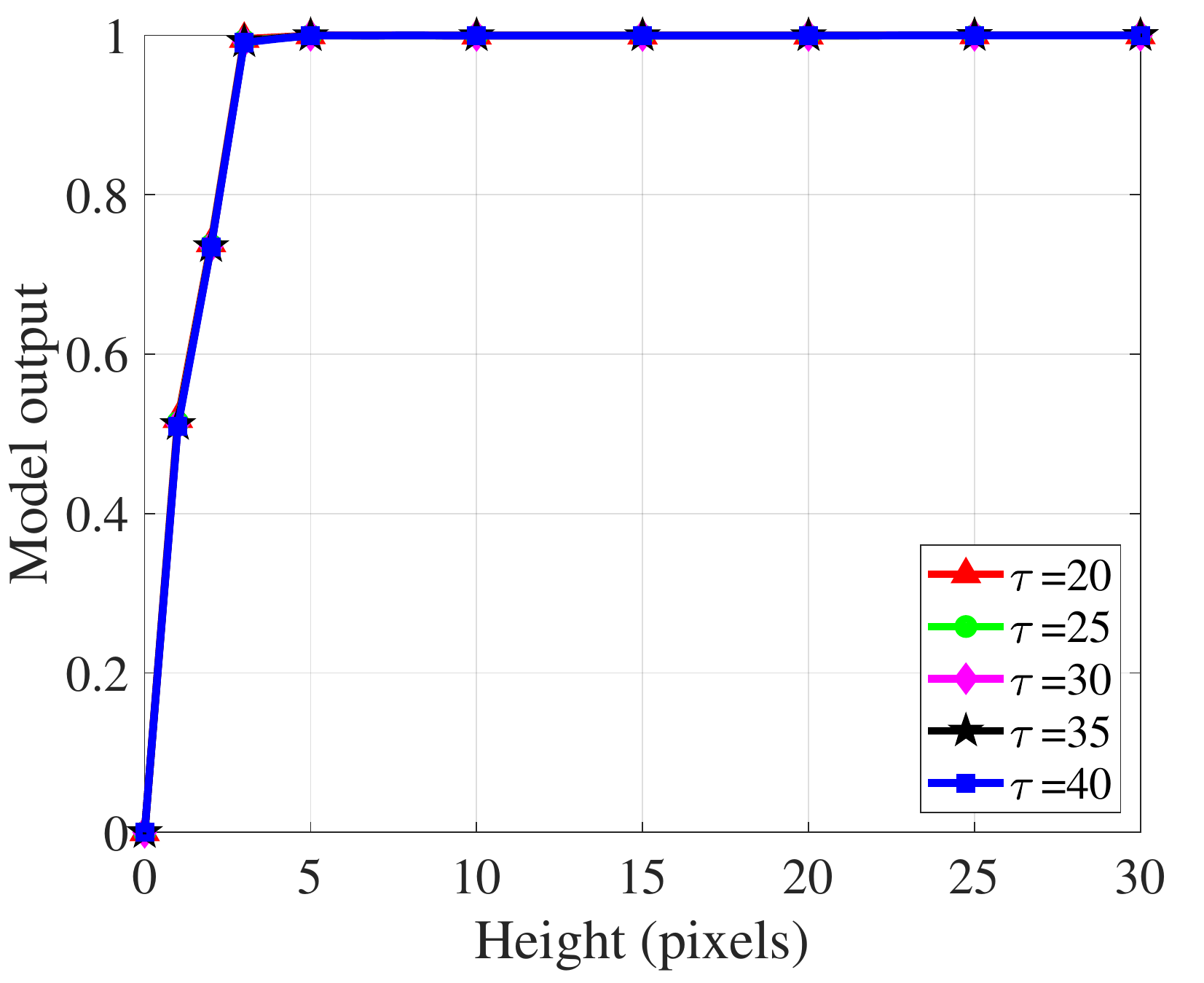}}
	\hfil
	\subfloat[]{\includegraphics[width=0.24\textwidth]{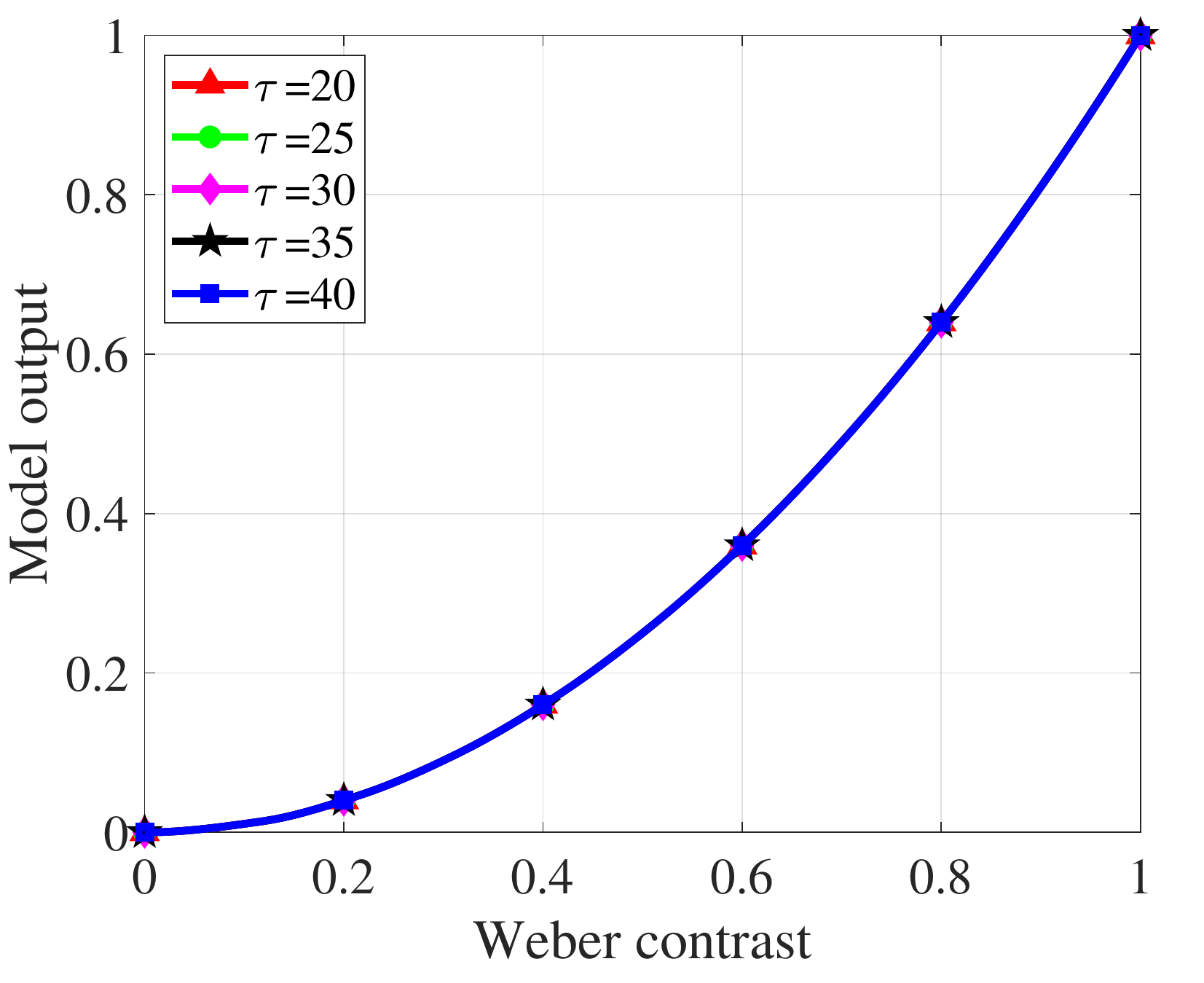}}
	\caption{Outputs of the LPTC neural model at various setting of the time constant $\tau$ regarding to (a) velocity, (b) width, (c) height, (d) Weber contrast of the object.}
	\label{EMD-Varying-Tau-Order-15-Marker}
\end{figure*}


\subsection{Parameter Sensitivity Study}
\label{Parameter-Sensitivity-Study}

Each LPTC neural model is determined by three parameters, including correlation distance $\beta$, order $n_5$, and time constant $\tau_5$, as can be seen from (\ref{Formulation-Two-Pixels-X-Y}) and (\ref{LPTC-Neuron-Correlation}). In this section, we evaluate the effects of these three parameters on the performance of the LPTC subnetwork.


We first study the effect of the correlation distance $\beta$ by tuning it in the range of $\{2, 6, 10, 14, 18\}$, while fixing $n_5 = 25$, $\tau_5 = 30$. Fig. \ref{Five-EMDs-Tuning-Curve-Marker}(a)-(d) illustrates the outputs of the LPTC model under various setting of $\beta$ regarding to (a) velocity, (b) width, (c) height, (d) Weber contrast of the object, respectively. Weber contrast measures the difference of average pixel intensity between the object and its surrounding area, whose formulation is given in the previous literature \cite{wang2021time}. As shown in Fig. \ref{Five-EMDs-Tuning-Curve-Marker}(a), for any given correlation distance $\beta$, the LPTC is highly selective to object velocity. Specifically, the LPTC exhibits positive responses in a specific velocity range (called preferred velocity range), and reach its maximal response at an optimal velocity. In addition, the velocity tuning curve changes significantly with respect to different settings of $\beta$. To be more precise, as $\beta$ is increased from $2$ to $18$ pixels, the optimal velocity rises from $150$ to $600$ pixels/s while the preferred velocity range is also gradually shifted toward higher velocities. As can be seen from Fig. \ref{Five-EMDs-Tuning-Curve-Marker}(b) and (c), the LPTC does not exhibit strong selectivity for object width and height. For a fixed $\beta$, the response of the LPTC remains stable to object width larger than $10$ pixels (or height larger than $5$ pixels) after it reaches the maximum at width $=10$ pixels (or height $=5$ pixels). In addition, the responses with respect to object width and height are little affected by the increase of $\beta$, though the responses to width lower than $10$ pixels shows a slight increase. From Fig. \ref{Five-EMDs-Tuning-Curve-Marker}(d), we can see that the LPTC shows Weber contrast sensitivity where the increase in Weber contrast leads to the increase in the model output, which finally reaches its strongest response at Weber contrast $= 1$. Furthermore, the Weber contrast sensitivity tuning curve remains unchanged at different settings of $\beta$.

We further study the performance variation of the LPTC model with respect to different values of the order $n_5$, where $n_5$ is tuned within $\{5, 10, 15, 20, 25\}$ while $\beta$ and  $\tau_5$ are fixed to $4$ and $30$, respectively. As shown in Fig. \ref{EMD-Varying-Order-Tau-30-Marker}(a), the decrease of the order $n_5$ results in the broader span of the preferred velocity while the optimal velocity remains almost unchanged. In Fig. \ref{EMD-Varying-Order-Tau-30-Marker}(b), the increase of the order $n_5$ has small inhibitory effect on the model outputs to object width lower than $10$ pixels. However, when object width is larger than $10$ pixels, changes in the model output resulting from the adjustment of $n_5$ become little. From Fig. \ref{EMD-Varying-Order-Tau-30-Marker}(c) and (d), we can observe that the order $n_5$ has minor influence on both height tuning curve and Weber contrast tuning curve. 

\begin{figure}[!t]
	\vspace{-1mm}
	\centering
	\includegraphics[width=0.275\textwidth]{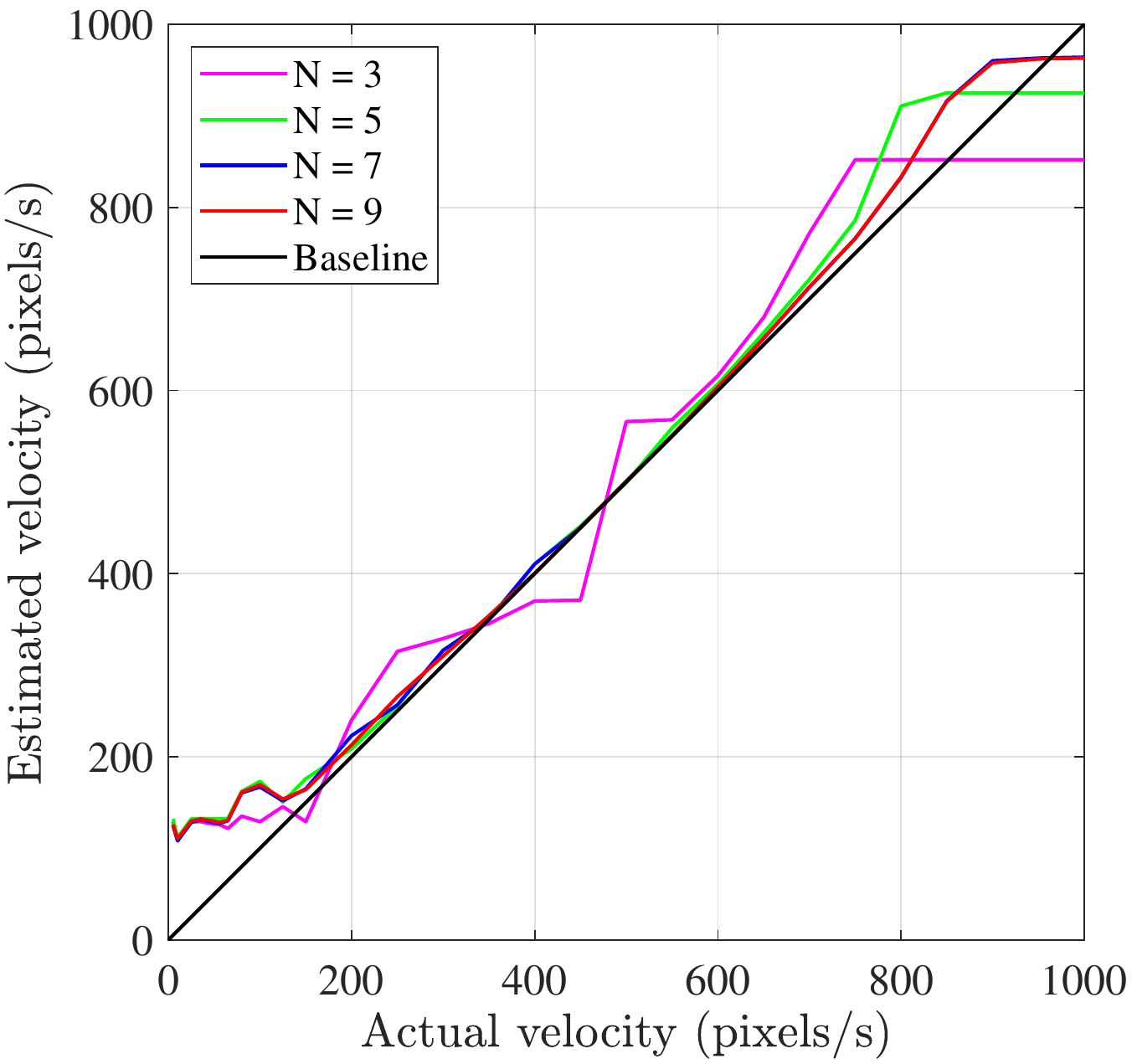}
	\caption{Object velocity estimated by the population coding mechanism at various setting of neuron number $N$. }
	\label{Velocity-Ground-Truth-Population-Coding}
\end{figure}

Finally, we investigate the performance changes of the LPTC under different setting of the time constant $\tau_5$ which is tuned within $\{20, 25, 30, 35, 40\}$ while $\beta$ and $n_5$ are fixed to $4$ and $15$, respectively. From Fig. \ref{EMD-Varying-Tau-Order-15-Marker}(a), it can be observed that the optimal velocity is slightly shifted toward higher values by the decrease in $\tau_5$. The preferred velocity range is also broadened, where the maximum of the preferred velocity range increases to a higher value about $800$ pixels/s when $\tau_5$ decreases to $20$, but the minimum remains unchanged at a low value about $50$ pixels/s. As shown in Fig. \ref{EMD-Varying-Tau-Order-15-Marker}(b)-(d), the three tuning curves, i.e., width tuning, height tuning, and Weber contrast tuning, are insensitive to the changes of $\tau_5$.

Based on the above sensitivity studies, we properly tune the correlation distance $\beta$ to obtain multiple LPTCs with overlapped preferred velocity ranges for neural population coding. In Fig. \ref{Velocity-Ground-Truth-Population-Coding}, we further reveal the relationship between velocity estimated by the population coding mechanism and actual velocity under different setting of the LPTC neural number $N$. As can be seen, the increase in the number of the LPTCs will provide a more accurate estimation of object velocity within $[150,750]$ pixels/s. The velocity estimation range of the population coding could be further broadened achieved by adding the LPTC neurons with preferred velocities higher than $750$ pixels/s and/or lower than $150$ pixels/s. In the experiments, we set the number of the LPTC neurons as $9$ in the population coding mechanism for object velocity estimation.

\subsection{Effectiveness of the Neural Population Coding}

\begin{figure}[t!]
	\centering
	\subfloat[]{\includegraphics[width=0.35\textwidth]{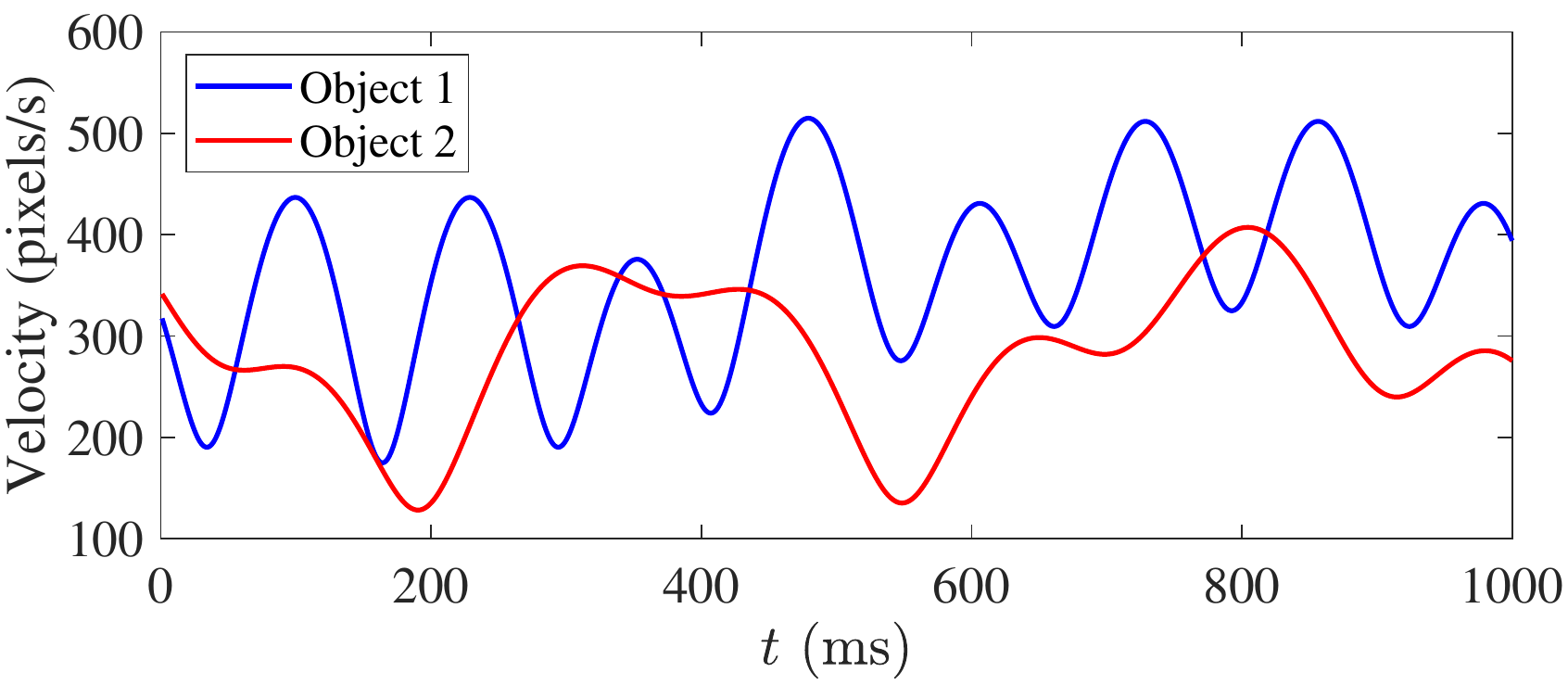}}
	\hfil
	\subfloat[]{\includegraphics[width=0.35\textwidth]{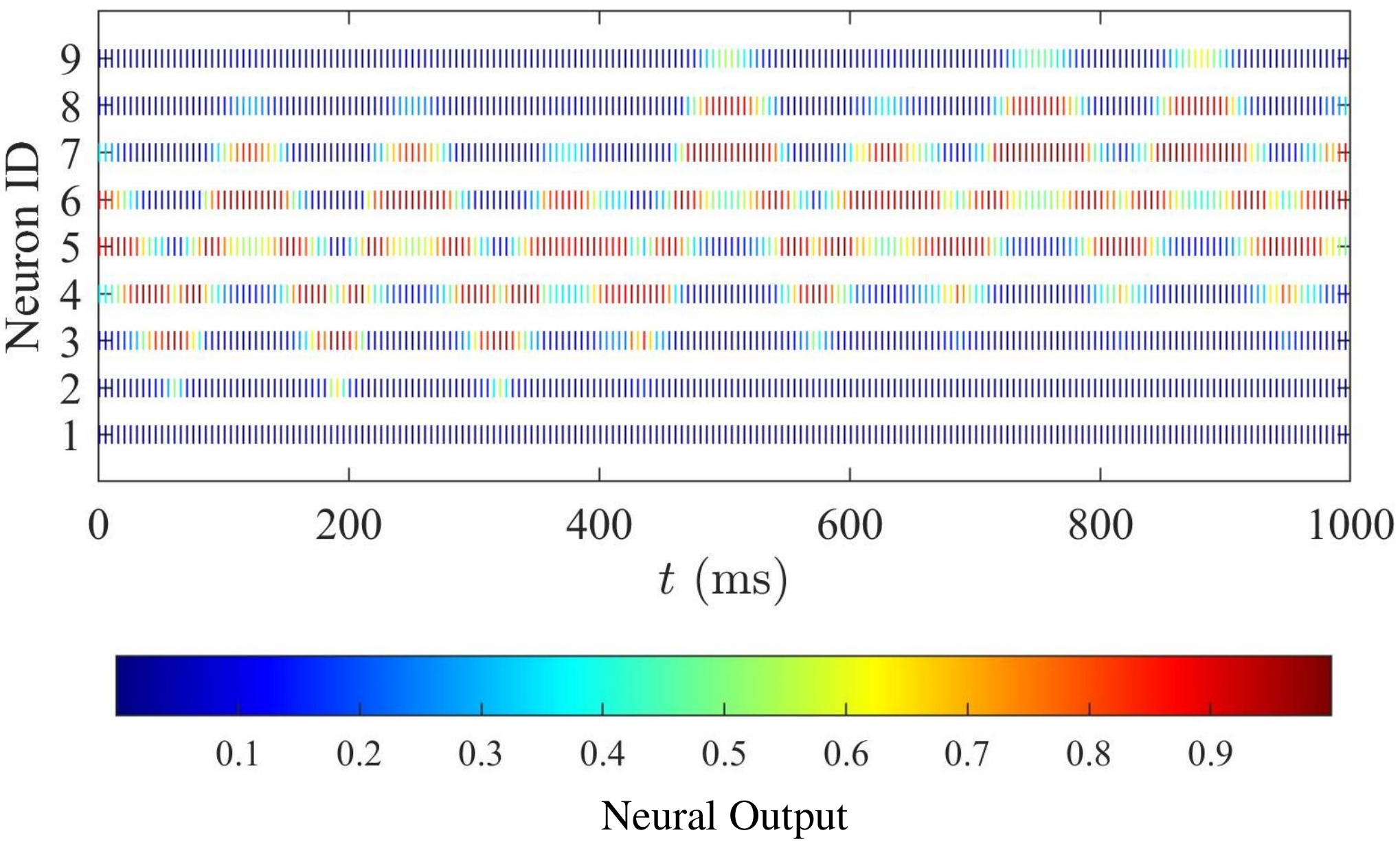}}
	\hfil
	\subfloat[]{\includegraphics[width=0.35\textwidth]{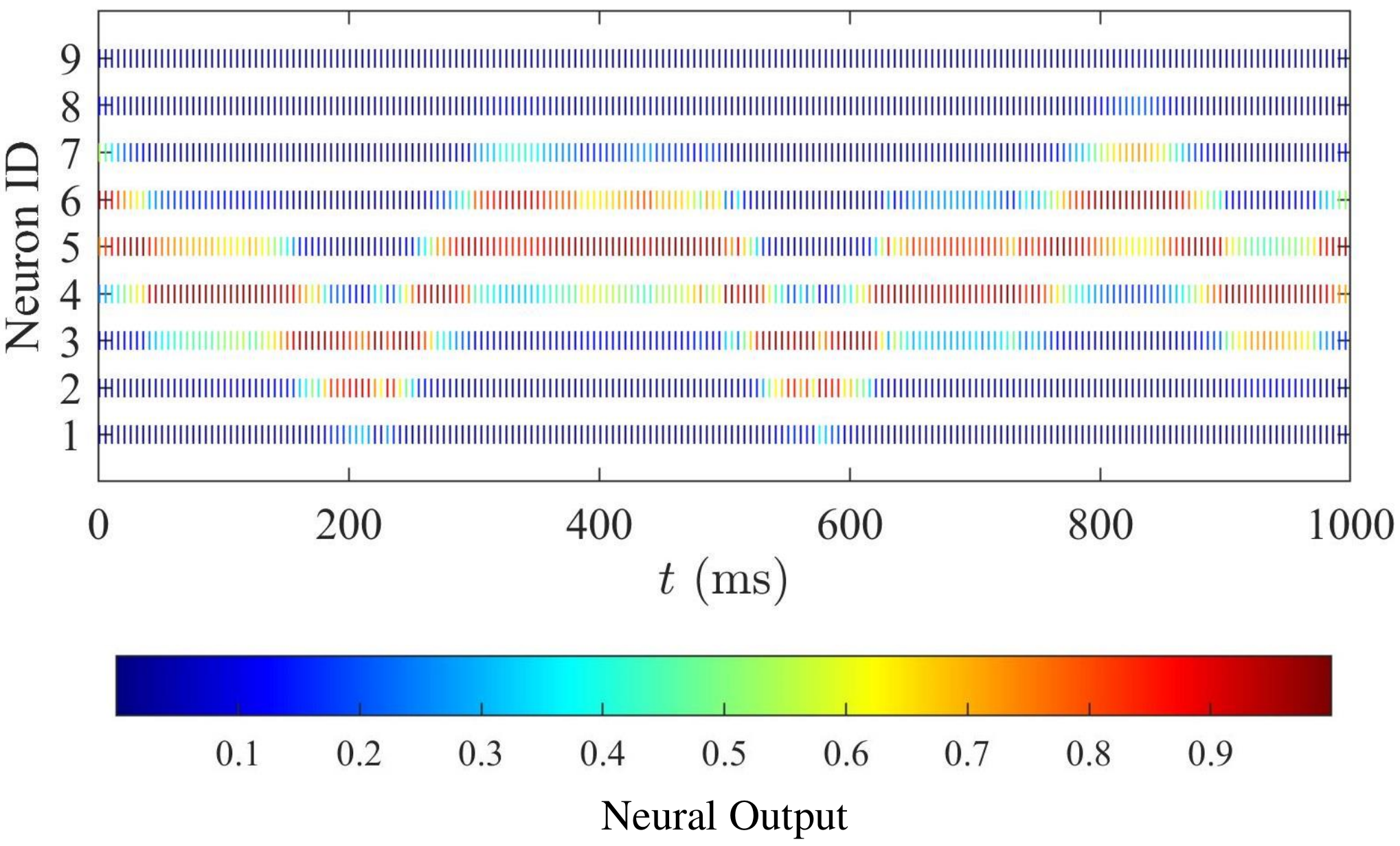}}
	\caption{(a) Velocities of two independently moving objects with respect to time $t$. (b) Raster plots of $N = 9$ LPTC neural outputs to the object $1$. (c) Raster plots of $N = 9$ LPTC neural outputs to the object $2$.}
	\label{Multi-EMDs-Responses}
\end{figure}

\begin{figure}[t!]
	\centering
	\subfloat[]{\includegraphics[width=0.40\textwidth]{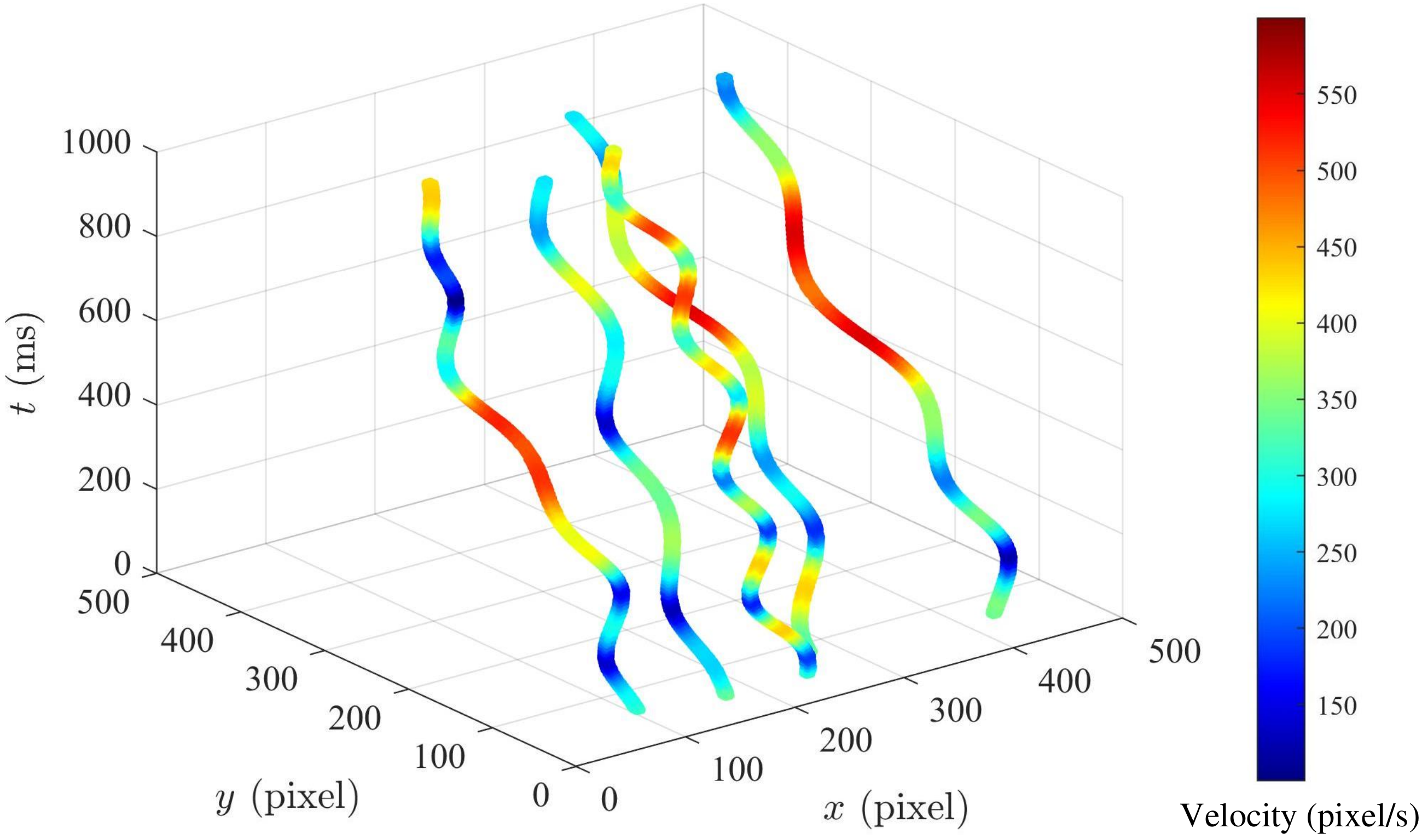}}
	\hfil
	\subfloat[]{\includegraphics[width=0.40\textwidth]{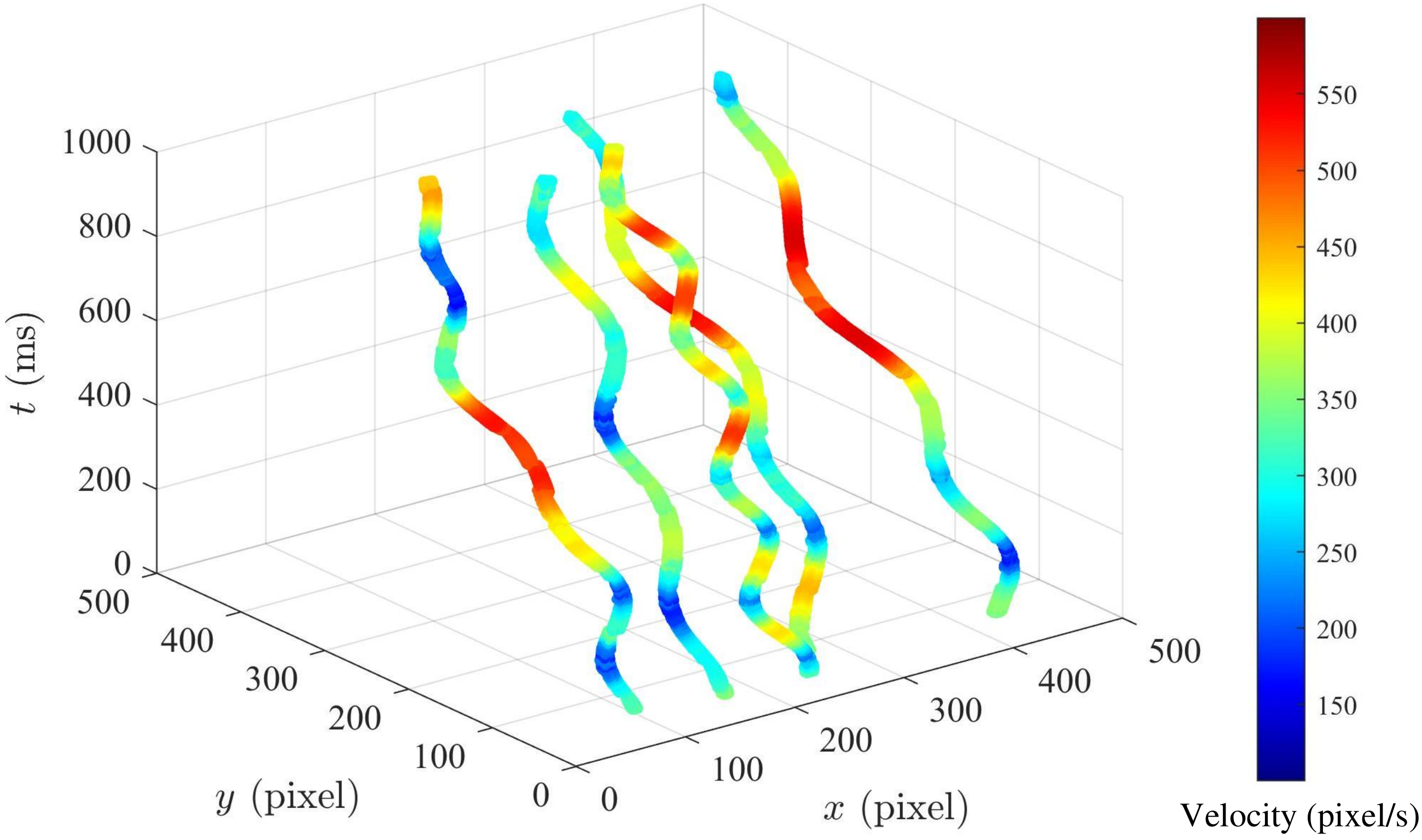}}
	\caption{(a) Actual trajectories and velocities, (b) trajectories and velocities estimated by the population coding mechanism in the three-dimensional (3D) view.}
	\label{Multi-EMDs-Velocity-Tensor-Estimation-Ground-Truth}
\end{figure}

We design a collection of the LPTC neurons with overlapping preferred velocity ranges to encode object velocity into neural activities. To validate its effectiveness, we conduct experiments on image sequences that hold multiple object motion with variable velocities. Fig. \ref{Multi-EMDs-Responses}(a) illustrates  velocities of two independently moving objects during time period $[0,1000]$ ms, where that of object $1$ fluctuates frequently and significantly while that of object $2$ exhibits a much more smooth change. The neural outputs of the LPTCs to these two objects over time are shown in Fig. \ref{Multi-EMDs-Responses}(b) and (c), respectively. It can be observed that the outputs of the LPTCs are represented by colored rasters, where the redder the raster, the stronger response of the corresponding neuron. A LPTC neuron will strongly fire only when object velocity is within its preferred velocity range; otherwise, it will fire sporadically or even not exhibit response. The larger neural ID means the corresponding LPTC neuron has a higher velocity range. It can also be  observed that the firing patterns of all the LPTC neurons for a moving object appears to encode the object velocity profile. Specifically, the ID of firing neurons are highly relevant to the object's velocity at a given time. In addition, neural responses are quite reliable across the nine LPTCs for each moving object.

Fig. \ref{Multi-EMDs-Velocity-Tensor-Estimation-Ground-Truth}(a) shows actual trajectories and velocities of five independently moving objects where velocity is represented by color. For comparison purpose, trajectories and velocities estimated by the proposed population coding mechanism are shown in Fig. \ref{Multi-EMDs-Velocity-Tensor-Estimation-Ground-Truth}(b). We can observe that the population coding mechanism provides a good estimation for object velocities. Despite the trajectories having different velocities, accelerations, and angles, the estimation still can roughly match the ground truth given in Fig. \ref{Multi-EMDs-Velocity-Tensor-Estimation-Ground-Truth}(a). The above results demonstrate the effectiveness of neural population coding in velocity estimation, which plays a role in the LPTC subnetwork to extract spatio-temporal dynamics of background movement.

\subsection{Effectiveness of the Spatio-Temporal Feedback}

\begin{figure}[!t]
	\centering
	\includegraphics[width=0.4\textwidth]{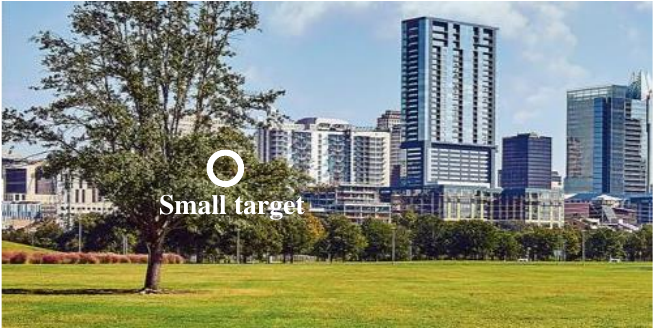}
	\caption{Representative image that contains a small moving object without significant visual features while exhibiting extremely low contrast to complex environment. The size of the small object roughly equates to $5 \times 5$ pixels while its velocity approximates to $350$ pixels/s which is lower than that of its surrounding background ($450$ pixels/s).}
	\label{Neural-Output-Input-Image}
\end{figure}

\begin{figure}[!t]
	\centering
	\includegraphics[width=0.48\textwidth]{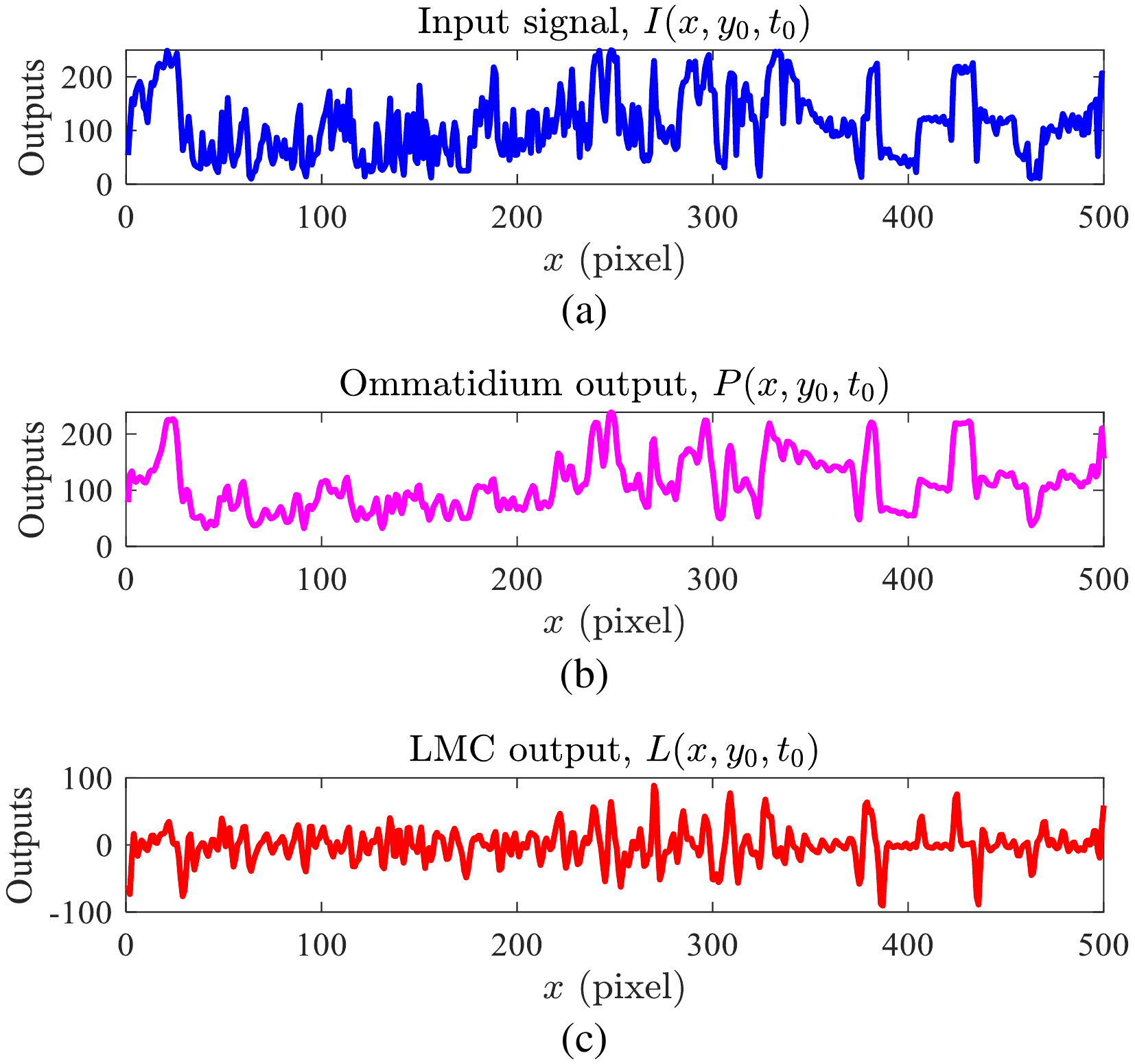}
	\caption{(a) Input signal $I(x,y_0,t_0)$ with respect to $x$ while fixing $y_0=125$ pixels and $t_0=500$ ms. (b) Ommatidium output of the retina layer $P(x,y_0,t_0)$. (c) LMC output of the lamina layer $L(x,y_0,t_0)$.}
	\label{Layer-Output-Input-Ommatidium-LMC}
\end{figure}

\begin{figure*}[!t]
	\centering
	\includegraphics[width=1\textwidth]{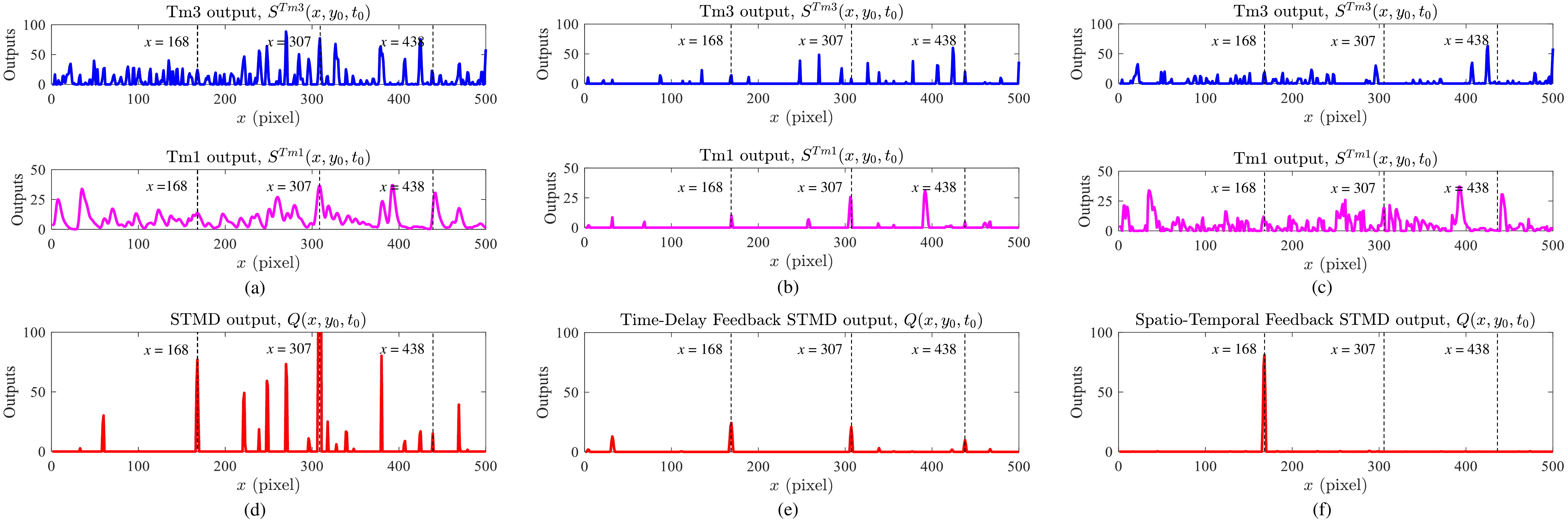}
	\caption{Top: Outputs of the two medulla neurons (a) without feedback, (b) with the time-delay feedback, and (c) with the spatio-temporal feedback.  Bottom: Outputs of the STMD neuron (d) without feedback, (e) with the time-delay feedback, and (f) with the spatio-temporal feedback.}
	\label{Layer-Output-Medulla-STMD-Feedback-STMD}
\end{figure*}

We develop a spatio-temporal feedback mechanism to suppress background false positives while enhancing responses to small moving targets. To validate its effectiveness, we compare the detection performances between the neural networks without feedback, with time-delay feedback, and with spatio-temporal feedback. Fig. \ref{Neural-Output-Input-Image} shows a representative frame of the input video to the three neural networks at time $t = 500$ ms. In the scenario, a small target, lacking discriminative visual features and presenting extremely low contrast, is moving amidst clutters of natural environment. Its velocity roughly equates to $350$ pixels/s whereas that of the surrounding background is about $450$ pixels/s. For better observation of signal processing in each neural layer, we fix $y_0 = 125$ and present neural outputs in relation to $x$, where the input signal $I(x,y_0,t_0)$, the ommatidium output of the retina layer $P(x,y_0,t_0)$, and the LMC output of lamina layer $L(x,y_0,t_0)$, are displayed in Fig. \ref{Layer-Output-Input-Ommatidium-LMC}(a)-(c), respectively. Because the time-delay feedback and spatio-temporal feedback are all implemented in the medulla and lobula layers, the three neural networks share the same ommatidium and LMC outputs. As shown, the ommatidium acts as a Gaussian blur achieving smoothing effects on the input signal while the LMC serves as a temporal filter to calculate luminance changes of each pixel over time. The positive and negative outputs of the LMC reflect luminance increase and decrease regarding to time, respectively.

Fig. \ref{Layer-Output-Medulla-STMD-Feedback-STMD}(a)-(c) presents medulla neural outputs without feedback, with the time-delay feedback, and with the spatio-temporal feedback, respectively. From Fig. \ref{Layer-Output-Medulla-STMD-Feedback-STMD}(a), we can find that the outputs of the two medulla neurons without feedback are derived from positive component and temporally-delayed negative component of the LMC output, respectively. They are multiplied together and then laterally inhibited to define the output of the STMD neuron. As shown in Fig. \ref{Layer-Output-Medulla-STMD-Feedback-STMD}(d), the STMD without feedback responds strongly to the small target at $x=128$, but it also exhibits significant outputs to background clutters, such as $x=307$ and $x=438$, due to mistaken correlation of luminance-change signals indued by background motion. To suppress these false-positive background responses, the STMD neural output is temporally delayed and further propagated to the lower layer as feedback signal for subtraction from the medulla neural outputs. It can be observed from Fig. \ref{Layer-Output-Medulla-STMD-Feedback-STMD}(b) that the outputs of the medulla neurons to the small target and background clutters are all dramatically weakened after the time-delay feedback, eventually leading to the decrease of the STMD neural outputs in Fig. \ref{Layer-Output-Medulla-STMD-Feedback-STMD}(e). However, the time-delay feedback cannot completely eliminate background false positives with high velocities (see $x=307$ and $x=438$). To overcome the limitation of the time-delay feedback, the spatio-temporal feedback is designed by considering unique spatio-temporal characteristics of the background motion. As shown in Fig. \ref{Layer-Output-Medulla-STMD-Feedback-STMD}(c), the two medulla neural outputs to background features are not only greatly suppressed but also properly aligned in time domain after the spatio-temporal feedback. These properly-aligned medulla neural outputs avoid mistaken correlation and finally generate the noiseless STMD neural output. As illustrated in Fig. \ref{Layer-Output-Medulla-STMD-Feedback-STMD}(f), the STMD with the spatio-temporal feedback filter out all background false positives while slightly enhancing response to the small target. The above experiment demonstrated the ability of the spatio-temporal feedback to suppress background false positives regardless of their velocities.

\subsection{Comparative Results on Synthetic and Real Data}
We conduct quantitative evaluation on the synthetic and real-world datasets, i.e., Vision Egg \cite{straw2008vision} and RIST \cite{RIST-Data-Set}, in the matter of three key metrics, i.e., detection rate, false alarm rate, and ROC curve. Specifically, we first carry out experiments on five groups of synthetic videos each of which holds a single small target motion but differs in three object parameters (i.e., size, luminance, and velocity) and two background parameters (i.e., velocity and motion direction), to investigate their effects on detection performance. Secondly, we validate the effectiveness of the proposed model for detecting multiple small moving targets against various backgrounds. Finally, we perform the evaluation on the challenging real-world data set. Three state-of-the-art models, including DSTMD \cite{wang2018directionally}, ESTMD \cite{wiederman2008model}, and Time-Delay Feedback STMD \cite{wang2021time}, are chosen for comparison, where their parameter settings are same with those in \cite{wang2018directionally,wiederman2008model,wang2021time}. 

\begin{figure*}[t!]
	\vspace{-2.5mm}
	\centering
	\includegraphics[width=0.90\textwidth]{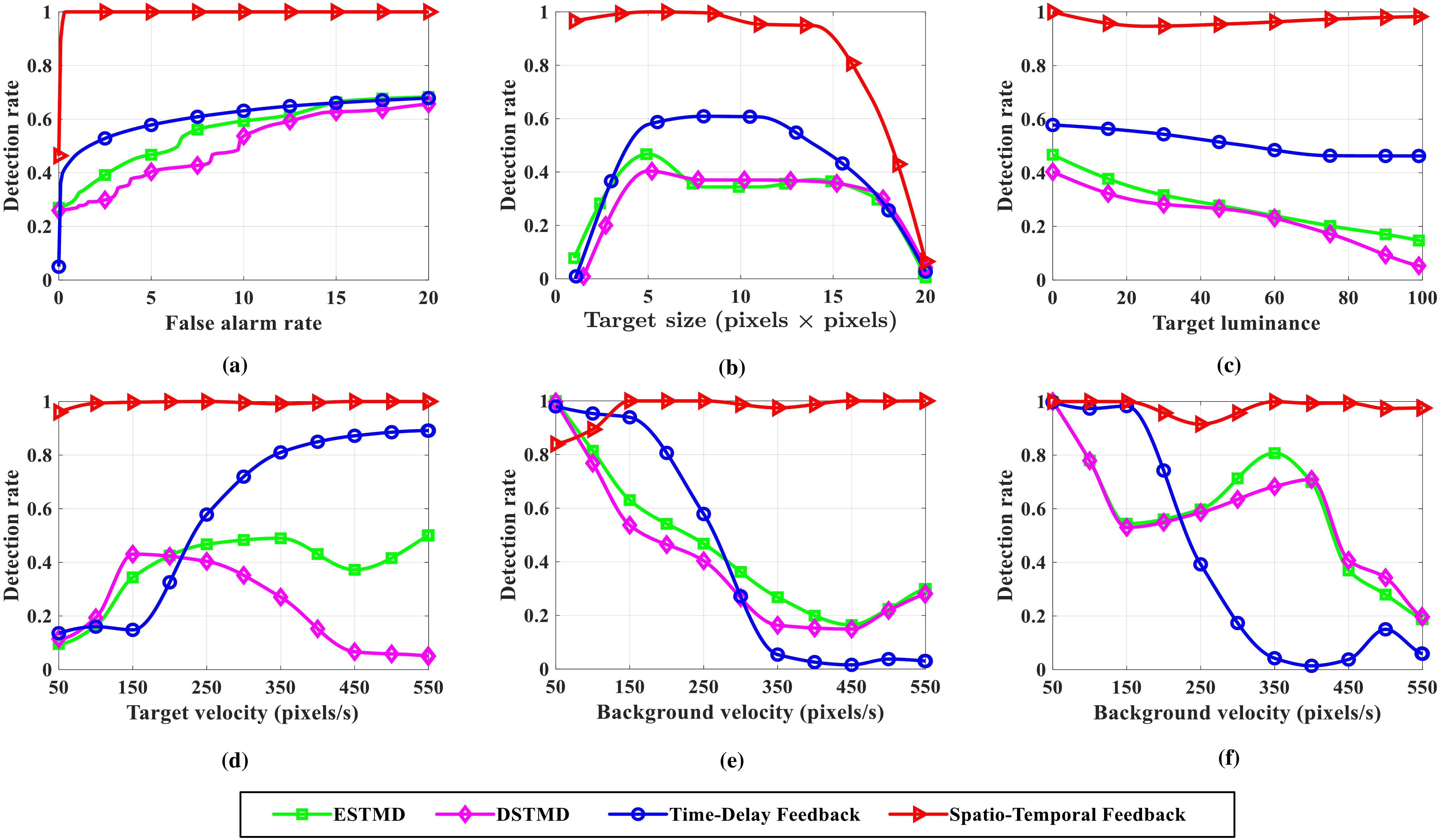}
	\caption{(a) ROC curves of the four models on the initial video. (b)-(f) Detection rates of the four models when false alarm rate $F_A = 5$, in relation to (b) target size, (c) target luminance, (d) target velocity, (e) background velocity (leftward motion), and (f) background velocity (rightward motion).}
	\label{CB-1-ROC-Varying-Paras}
\end{figure*}

\subsubsection{Performances on Videos with Varying Parameters}
The five parameters of the initial video, i.e., target size, target luminance, target velocity,  background velocity, and background motion direction, are
set as $5 \times 5$ pixels, $25$, $250$ pixels/s, $250$ pixels/s, and leftward, respectively. Each group of synthetic videos tunes one of the parameters while fixing others at their initial values. The comparison of the ROC curves on the initial video is present in Fig. \ref{CB-1-ROC-Varying-Paras}(a). It can be observed that the proposed spatio-temporal feedback model achieves the best performance. Specifically, it consistently has the highest detection rate close to $1$ at a fixed false alarm rate, in comparison to other three baseline models. 

In Fig. \ref{CB-1-ROC-Varying-Paras}(b)-(f), we further compare the detection rates of the four models when false alarm rate $F_A = 5$, in relation to the five parameters, respectively. As illustrated in Fig. \ref{CB-1-ROC-Varying-Paras}(b), the spatio-temporal feedback leads to significant performance improvements for all target sizes. In particular, its detection rate is close or even equal to $1$ when target size ranges from $1\times 1$ to $15 \times 15$ pixels, though experiences a sharp decrease for target size larger than $15 \times 15$ pixels. In contrast, the three competing models have a preferred size range roughly between $5\times 5$ and $12 \times 12$ pixels where their maximal detection rates are much lower than that of the spatio-temporal feedback model. The result reveals that the spatio-temporal feedback is able to alleviate performance degradation induced by the size selectivity, but still cannot detect large objects with sizes larger than $20 \times 20$ pixels. From Fig. \ref{CB-1-ROC-Varying-Paras}(c), we can see that the spatio-temporal feedback model achieves the best performance over the competing methods under various levels of luminance. Specifically, the spatio-temporal feedback maintains extremely high detection rate (close to $1$) for target luminance ranging between $0$ and $100$. In contrast, the highest detection rates of the competing methods are all peaked at target luminance $0$ and lower than $0.6$. In addition, their detection performances substantially degrade with the increase in target luminance. The result indicates that the spatio-temporal feedback is able to alleviate the strong dependence on visual contrast between the small target and its surrounding background. 

As shown in Fig. \ref{CB-1-ROC-Varying-Paras}(d)-(f), the spatio-temporal feedback model clearly outperforms the competing models under various target and background velocities. Its detection rate remains stable (close to $1$) in most cases, while experiencing a slight decrease for background velocity lower than $150$ pixels/s in leftward motion. Although the time-delay feedback improves the model performance in detecting those objects moving faster than the surroundings, but it exhibits a significant performance degradation and performs the worst among the four models when the velocity of small target is lower than that of the background. The above results demonstrate that the spatio-temporal feedback leads to significant performance improvements in almost all cases if there is velocity difference between small target and its surrounding background.

\subsubsection{Performances on Videos with Multiple Small Targets in Various Backgrounds}
\begin{figure*}[t!]
	\centering
	\includegraphics[width=0.90\textwidth]{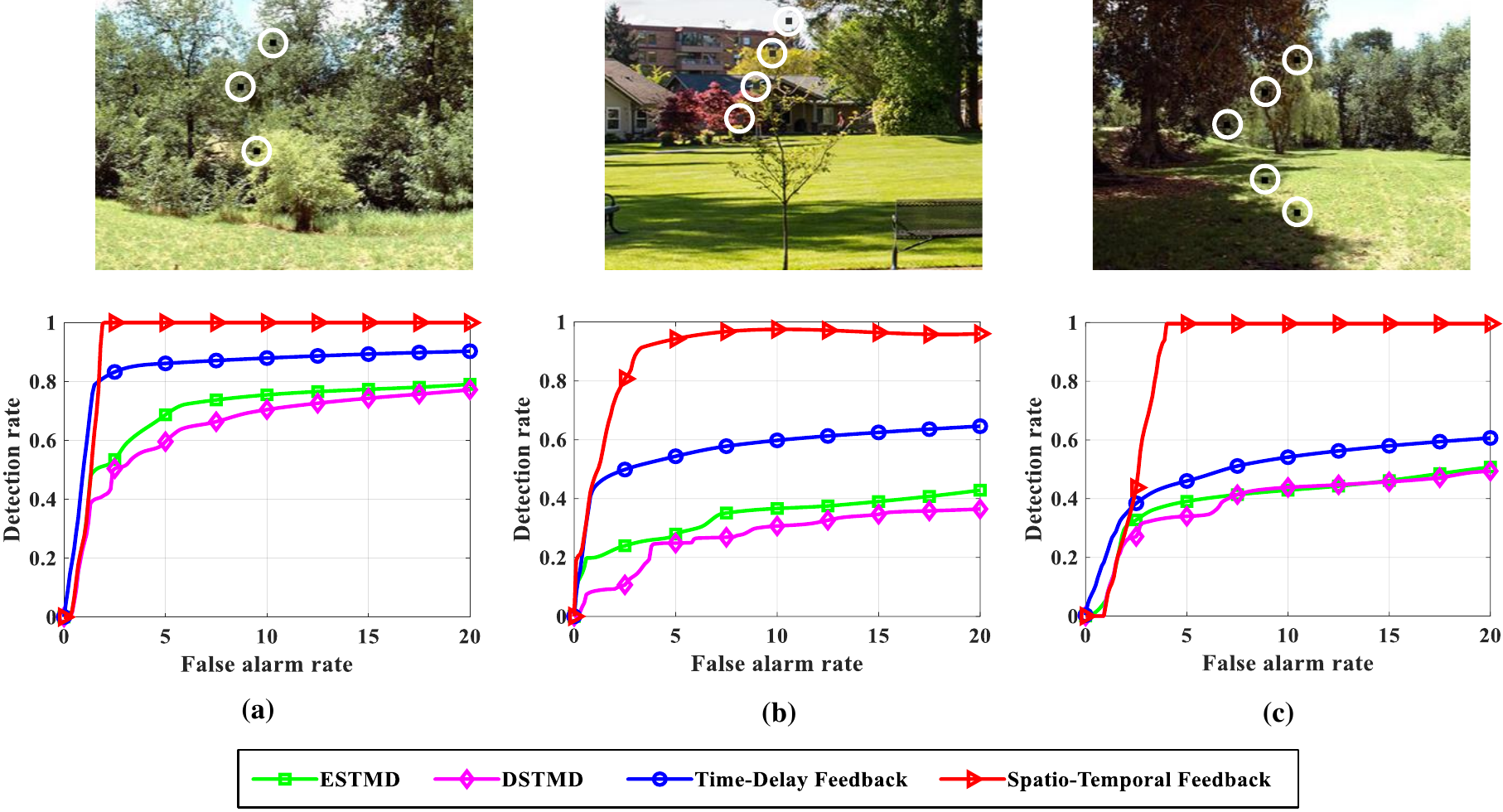}
	\caption{(a)-(c) Representative frames (top) and ROC curves of the four models (bottom) on three videos each of which presents multiple small target motion in various natural backgrounds.}
	\label{ROC-Curves-Synthetic-Videos-Diff-Background}
\end{figure*}

The ROC curve comparison of the four models on synthetic videos with three, four, and five small targets, are shown in Fig. \ref{ROC-Curves-Synthetic-Videos-Diff-Background}(a)-(c), respectively. We can observe that the spatio-temporal feedback achieves a substantial improvement over the other three methods for detecting multiple small targets against various complex backgrounds. It has an obvious advantage over the Time-Delay Feedback, DSTMD, and ESTMD models under any given false alarm rate. For example, the detection rate $D_R$ of the spatio-temporal feedback reaches $1$ when false alarm rate $F_A$ is set as $5$, representing improvements by a margin of $14$ percent over the Time-Delay Feedback,  $32$ percent over the ESTMD, and $40$ percent over the DSTMD, as illustrated in Fig. \ref{ROC-Curves-Synthetic-Videos-Diff-Background}(a).

\subsubsection{Performances on Real-World Videos}
\begin{figure*}[t!]
	\centering
	\includegraphics[width=0.90\textwidth]{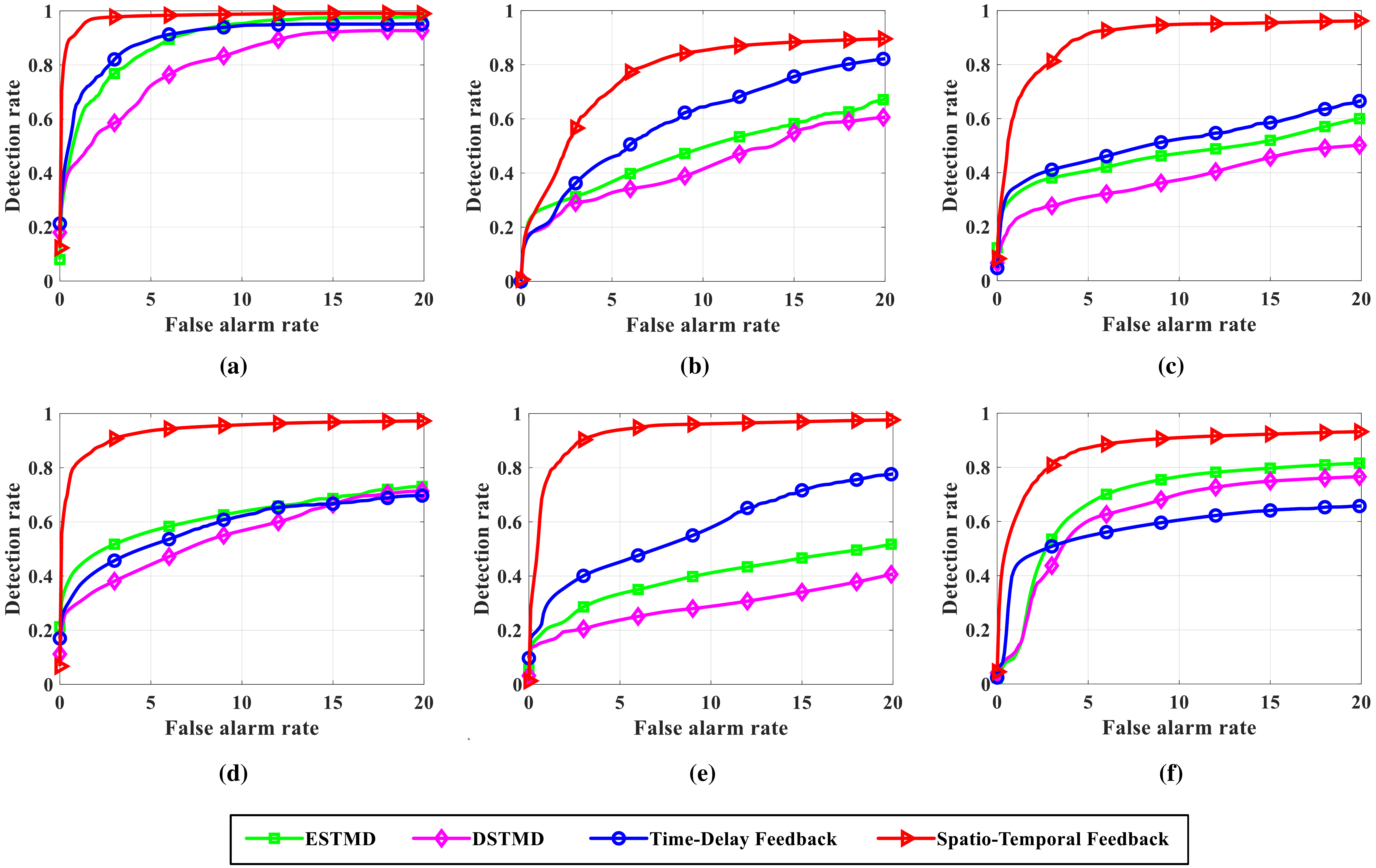}
	\caption{ROC curves of the for models on six real-world image sequences in the RIST data set, i.e., (a) GX010220, (b) GX010241, (c) GX010285, (d) GX010290, (e) GX010300, and (f) GX010315.}
	\label{ROC-Curves-Real-Videos}
\end{figure*}

We report the comparative results of the four methods in terms of the ROC curves on six real-world videos randomly selected from the RIST data set \cite{RIST-Data-Set}. The numbers of the selected videos are GX010220, GX010241, GX010285, GX010290, GX010300, and GX010315, respectively, each of which presents a small target motion in complex dynamic environment. As shown in Fig. \ref{ROC-Curves-Real-Videos}(a)-(f), the spatio-temporal feedback model achieves the best performance on all six real-world videos. Its detection rates are clearly higher than those of the competing methods by a large margin under any fixed false alarm rate. The results demonstrate that the proposed spatio-temporal feedback mechanism can effectively improve the model robustness in complex real environments.

\section{Conclusion}
\label{Conclusion}
We have presented a spatio-temporal feedback neural network for discriminating small target motion amidst heavy clutters of natural environment. The key difference from prior works lies in that we incorporate spatial and temporal dynamics of background motion into feedback signal to suppress false positives in an recurrent manner. To this end, two subnetworks called LPTC and STMD, are proposed to extract motion information of the cluttered background and small targets, respectively. The spatio-temporal dynamics of the background is relayed from the LPTC to the STMD via an inter-layer connection, and further integrated with the output of the STMD as a feedback signal. The elimination of false positives is achieved by mixing the inputs to the STMD with the negative feedback signal, which forming a self-feedback loop. With the designed spatio-temporal feedback, the network performance in the task of small target motion detection is greatly improved.


\ifCLASSOPTIONcaptionsoff
  \newpage
\fi


\bibliographystyle{IEEEtran}

\bibliography{IEEEabrv,Reference}

\end{document}